\documentclass[a4paper]{article}
\usepackage[utf8]{inputenc}
\usepackage{float}
\usepackage{caption}
\usepackage[margin=1in]{geometry}
\usepackage{subcaption}
\usepackage{multirow}
\usepackage{cite}
\usepackage{amsmath}
\usepackage{amsfonts}
\usepackage[toc,page]{appendix}
\usepackage{graphicx}
\usepackage{nicefrac}

\title{\bf{A tomographic workflow to enable deep learning \\for X-ray based foreign object detection}}

\author{
  Math\'e T. Zeegers$^{1}$\thanks{Corresponding author: m.t.zeegers@cwi.nl, Science Park 123, 1098 XG Amsterdam, The Netherlands} \and Tristan van Leeuwen$^{1,2}$ \and Dani\"el M. Pelt$^{1,3}$ \and Sophia Bethany Coban$^{1,4}$ \and Robert van Liere$^{1,5}$ \and Kees Joost Batenburg$^{1,3}$
}

\date{{\small
    $^{1}$ Centrum Wiskunde \& Informatica,  Science Park 123, 1098 XG Amsterdam, The Netherlands \\
    $^{2}$ Mathematical Institute, Utrecht University, Budapestlaan 6, 3584 CD Utrecht, The Netherlands\\
    $^{3}$ Leiden Institute of Advanced Computer Science, Niels Bohrweg 1, 2333 CA Leiden, The Netherlands\\
    $^{4}$ Department of Mathematics, University of Manchester, Oxford Road, Manchester, M13 9PL. United Kingdom \\
    $^{5}$ Faculteit Wiskunde en Informatica, Technical University Eindhoven, Groene Loper 5, 5612 AZ Eindhoven, The Netherlands}
    {\normalsize \ \\ \today}
}

\begin{document}

\maketitle

\begin{abstract}
    Detection of unwanted (\lq foreign') objects within products is a common procedure in many branches of industry for maintaining production quality. X-ray imaging is a fast, non-invasive and widely applicable method for foreign object detection. Deep learning has recently emerged as a powerful approach for recognizing patterns in radiographs (i.e., X-ray images), enabling automated X-ray based foreign object detection. However, these methods require a large number of training examples and manual annotation of these examples is a subjective and laborious task. In this work, we propose a Computed Tomography (CT) based method for producing training data for supervised learning of foreign object detection, with minimal labour requirements. In our approach, a few representative objects are CT scanned and reconstructed in 3D. The radiographs that have been acquired as part of the CT-scan data serve as input for the machine learning method. High-quality ground truth locations of the foreign objects are obtained through accurate 3D reconstructions and segmentations. Using these segmented volumes, corresponding 2D segmentations are obtained by creating virtual projections. We outline the benefits of objectively and reproducibly generating training data in this way compared to conventional radiograph annotation. In addition, we show how the accuracy depends on the number of objects used for the CT reconstructions. The results show that in this workflow generally only a relatively small number of representative objects (i.e., fewer than $10$) are needed to achieve adequate detection performance in an industrial setting. Moreover, for real experimental data we show that the workflow leads to higher foreign object detection accuracies than with standard radiograph annotation.
\end{abstract}


\section{Introduction}
Foreign object detection in an industrial high-throughput setting is essential for guaranteeing quality and safety of objects processed in factory lines. Foreign objects may, for example, appear in products such as meat, fish or vegetables as small pieces of glass, bones, plastic, wood or stone that could harm consumers \cite{AndriiashenLiere, Wilm, ZhuSpachos}. Conventional nondestructive methods for detecting foreign objects include ultrasound imaging, X-ray imaging, magnetic resonance imaging, fluorescence imaging, (hyperspectral) spectroscopic imaging and thermal imaging \cite{HeXiao, LiLuo, KhairiIbrahim, NarsaiahBiswas, NicolaiDefraeye, XiongSun}. X-ray imaging provides the unique opportunity to visualize the interior structure of an object in a fast, low-cost, and non-invasive manner. This enables \emph{X-ray based foreign object detection}, in which the goal is to detect unwanted smaller objects inside base objects based on their distinct attenuation or attenuation patterns, as observed in generated \emph{radiographs} (i.e. standard 2D X-ray images). The possibility to reveal hidden foreign objects on radiographs has lead to its extensive use in various industrial applications \cite{NarsaiahBiswas, EinarsdottirEmerson, HaffToyofuku, KwonLee, MathankerWeckler, MeryLillo, ZhongZhang}, for which low-cost, adaptive and efficient image processing methods are essential \cite{XiongSun, MathankerWeckler}. One way to achieve better discrimination of foreign objects in radiographs is to use multispectral X-ray imaging detectors, simultaneously capturing radiographs at two or more energy levels \cite{SiBar, TaguchiBlevis}. As the attenuation properties of each material have their own characteristic dependence on the X-ray energy, these multispectral images can be analyzed to extract material composition information. \\

However, superposition of materials gives rise to similar levels of intensities for different objects in 2D radiographs. This problem limits the application of commonly used segmentation methods, such as threshold-based, clustering-based, and boundary-based or edge-based segmentation \cite{SezginSankur, SilvaOliveira}, to extract different components of the object. Additionally, high-throughput acquisition may lead to high noise levels in radiographs, and this increases the difficulty of successful foreign object detection even further \cite{XiongSun, MathankerWeckler}. Commonly used segmentation methods can be unsuitable in case of poor image qualities caused by conditions such as noise, low contrast and homogeneity in regions close to foreign objects \cite{SilvaOliveira}. Most conventional unsupervised methods can therefore not achieve high accuracies \cite{SilvaOliveira} without extensive manual parameter tuning to use a method for a specific problem \cite{Al-SarayrehReis, RongXie}.\\

Machine learning is a powerful tool for recognizing patterns in images \cite{ZhaoZheng} and can potentially detect foreign objects in radiographs \cite{ZhuSpachos}. Recent machine learning methods address a wide variety of segmentation problems \cite{SilvaOliveira, GarciaEscolano}, and provide a remarkable improvement over more classical segmentation methods in many practical applications \cite{GuoLiu}. A key obstacle in the application of machine learning is the need for large datasets \cite{ChartrandCheng, WuLiu, DeshpandeMinai}, which is particularly prominent in machine learning for foreign object detection as each new combination of sample, foreign object, and imaging settings requires additional data. On top of that, supervised learning uses labeled datasets for training. However, manual annotation (as in e.g. \cite{SilvaOliveira, Al-SarayrehReis}) requires tremendous efforts \cite{AkcayToby}, is time consuming and tedious \cite{TajbakhshJeyaseelan}, is subjective and can be prone to errors.\\

\begin{figure}[!b]
    \centering
    \subcaptionbox{2D radiograph \label{fig:ProjectionDataLabObj3Intro}}[0.3\textwidth]{
		\includegraphics[width=0.3\textwidth]{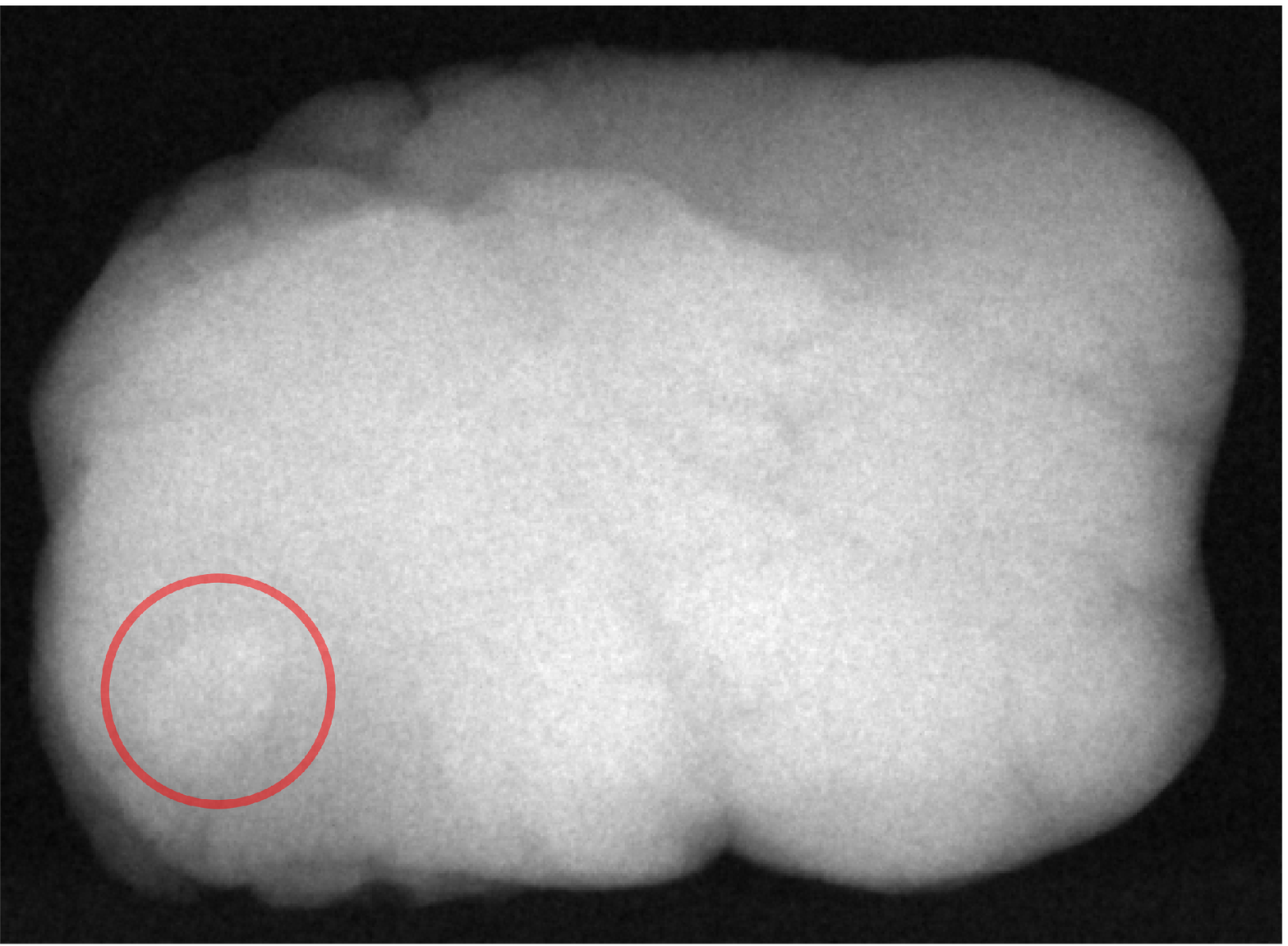}} \ \
	\subcaptionbox{2D slices in the reconstructed 3D volume space\label{fig:3DSlices}}[0.36\textwidth]{
		\includegraphics[width=0.36\textwidth]{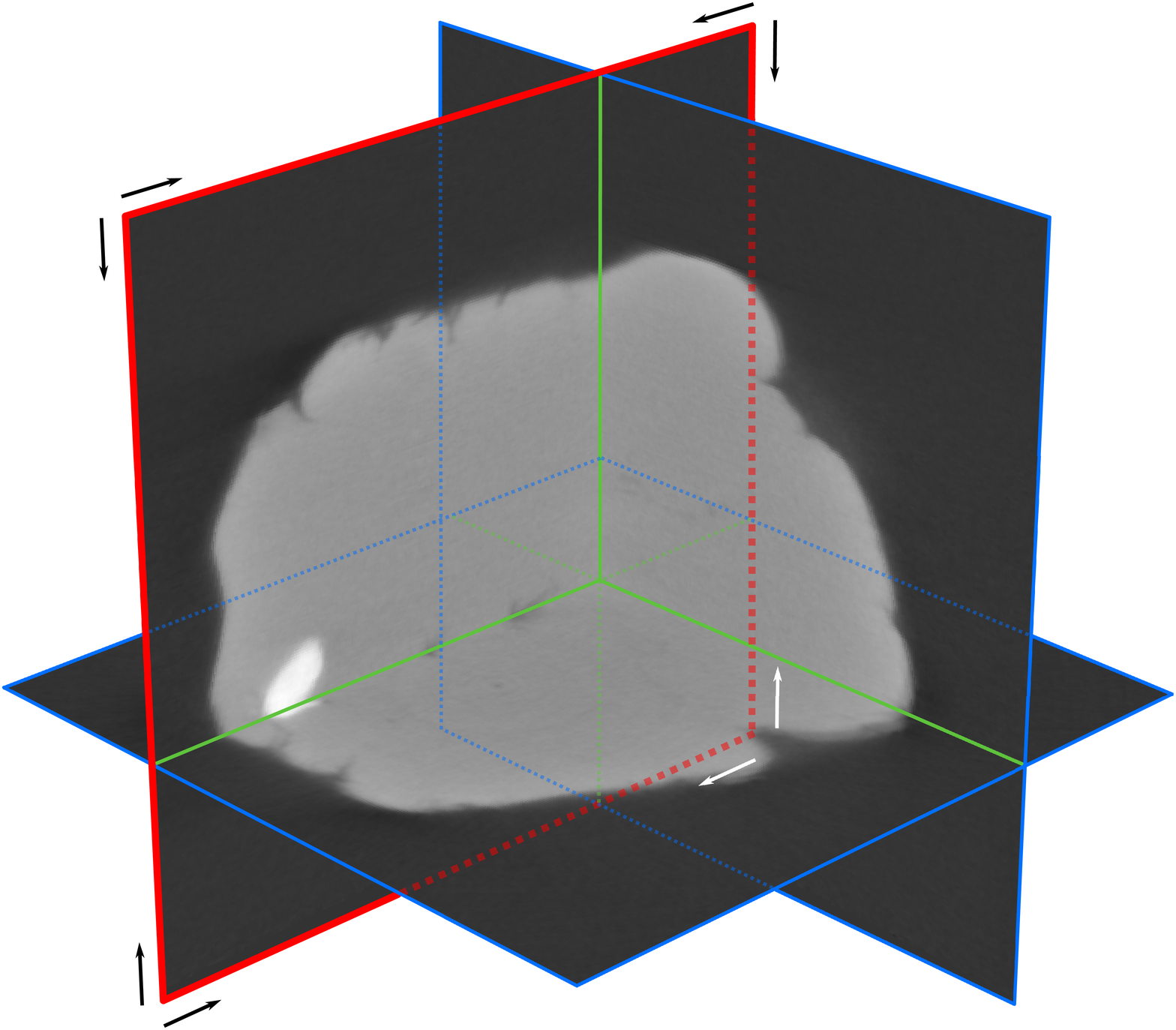}}
	\subcaptionbox{Red-bordered 2D slice of the 3D reconstruction\label{fig:ReconstructionObj3Intro}}[0.3\textwidth]{
		\includegraphics[width=0.3\textwidth]{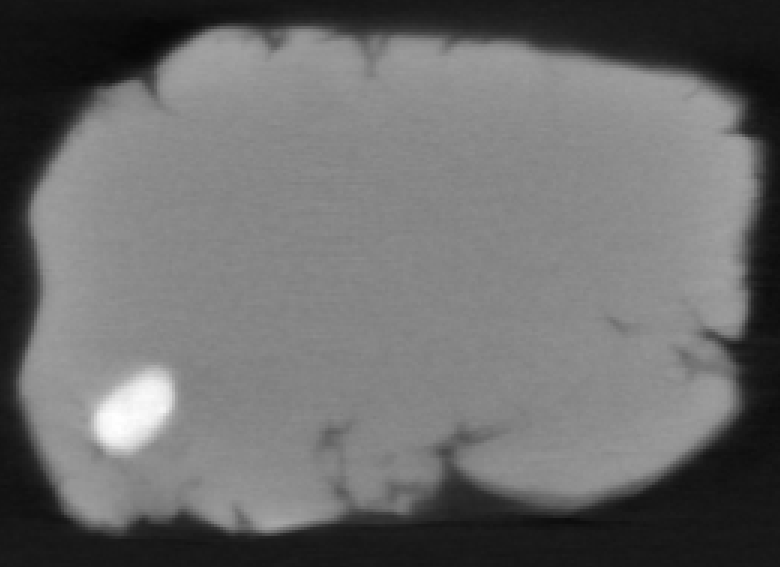}} \\
	\caption{Different views of an imaged product (Play-Doh) with a foreign object inserted (a piece of gravel). A 2D radiograph with the location of the foreign object (red circle) is shown (\textbf{a}), as well as multiple slices through the 3D volume of the reconstructed object (\textbf{b}), of which the slice with the red border is highlighted (\textbf{c}). The images show the difference in contrast: the foreign object is much easier to distinguish based on intensity values in the reconstructed 3D volume than in the 2D radiograph.}\label{fig:SeparationIntro}
\end{figure}

The key contribution of this paper is to propose a workflow based on 3D Computed Tomography (CT) for efficiently creating large training datasets, overcoming the aforementioned obstacle. CT scans of a relatively small number of objects are carried out with low exposure time -- as in a high-throughput setting -- yielding a large number of radiographs that are used as input for the supervised machine learning method. The same set of radiographs is also used offline for generating multiple high-quality tomographic 3D reconstructions, from which foreign objects can easily be segmented in 3D and projected back onto a virtual 2D detector to give the corresponding ground truth locations of the foreign objects on the radiographs. Without the effort of extensive manual labelling, this results in a large dataset with which deep learning can be carried out to detect foreign objects from fast-acquisition radiographs at a high rate. The example in Figure~\ref{fig:SeparationIntro} illustrates the difference in ease of segmentation for a CT reconstructed 3D volume versus a 2D radiograph. Whereas segmenting the foreign object in a radiograph is a challenging task, simple global thresholding can be applied to the CT volume to separate the foreign object from the base object. Additionally, more sophisticated and accurate segmentation and denoising rules can be imposed on 3D volumes \cite{GarciaEscolano, PanZhou, LooverboschRaeymaekers} than on 2D radiographs. \\

The structure of the paper is as follows. Section~\ref{section:Methods} provides the background of applying machine learning for foreign object detection, and explains the proposed method of data generation to apply machine learning. In Section~\ref{section:Experiments}, the workflow is demonstrated in a laboratory experiment, and shows how the number of imaged objects affects the detection accuracies. Additionally, the robustness of the workflow is analyzed. Section~\ref{section:Discussion} discusses various aspects of the results and the flexibility and modularity of the workflow. Section~\ref{section:Conclusions} presents the conclusions from this work.
    
\section{Methods} \label{section:Methods}

In this section, we introduce machine learning for foreign object detection and explain the methodology of our CT-based workflow for creating training data.

\subsection{Foreign object detection with X-ray imaging}

\begin{figure}[H]	
\centering
\includegraphics[width=0.99\textwidth]{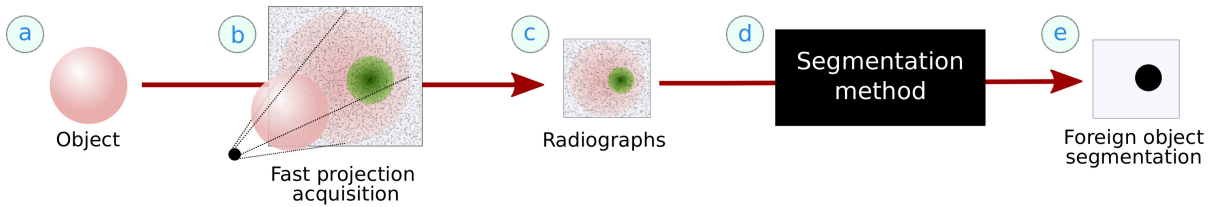}   \caption{A schematic overview of the foreign object detection problem and the segmentation-based approach to solving it. Each object (\textbf{a}) is assumed to have a correct segmentation (\textbf{e}). By using X-ray imaging (\textbf{b}), an X-ray radiograph of the object (\textbf{c}) can be acquired. Using a segmentation method (\textbf{d}), a segmented image (\textbf{e}) can be produced. The main challenge is to find a suitable segmentation method such that this approach to foreign object detection produces the correct results.}\label{fig:FODProblem}
\end{figure}

We consider the problem of foreign object detection in an industrial high-throughput conveyor belt setting. The problem and the usage of X-ray imaging to solve this are schematically shown in Figure~\ref{fig:FODProblem}. In foreign object detection, the aim is to correctly determine for each object whether a foreign object is contained in it or not, for instance a piece of bone within a meat sample.\\

For this problem, we focus on finding an accurate \emph{segmentation} for each radiograph. A segmentation partitions an image into sets of pixels with the same label. In our case, the formed segmented image is binary and indicates on which detector pixels a foreign object is projected. The segmentation depends on the type of objects that are considered to be foreign (by for instance a manufacturer). Any further classification (based on the minimum size of a foreign object for example) can be carried out after the segmented image is produced. \\

Throughout this paper, we use the term \emph{radiograph} for radiographs corrected using flatfield radiographs (without an object) and darkfield images (without the X-ray beam) that serve as input to the segmentation method. The quality of a radiograph depends on a number of properties of the scan, including exposure times, tube intensities, photon energy windows and the geometric setup \cite{Russo}. In a high-throughput setting, the steps in Figure~\ref{fig:FODProblem} should be fast to carry out, typically resulting in high noise levels and a challenging segmentation task.\\

\newpage

\subsection{Supervised learning} \label{sec:SupervisedLearning}
Machine learning is a widely used approach for difficult imaging tasks, as it can extract complicated patterns from complex images. In the foreign object detection problem, supervised machine learning can be used to learn the segmentation task such that it generalizes well for all possible fast-acquisition radiographs of similar objects with similar acquisition settings. To do so, a set of examples $\{(x_i,y_i)\}_{i=1}^{N}$ is used, where $\{x_i\}_{i=1}^{N}$ are acquired radiographs and $\{y_i\}_{i=1}^{N}$ are their corresponding foreign objects segmentations. The aim is to find the unknown segmentation function $F$ that maps each radiograph $x_i$ to its segmentation $y_i$. To find an approximate solution that generalizes well, the set of images is partitioned into a training set, a validation set and a test set. The training set is used to learn the function $F_{\text{train}}$ that minimizes the loss $L$ on the training set, which is the sum of errors between the segmented images $F_{\text{train}}(x_i)$ produced by the segmentation function and the true segmented images $y_i$. To find a suitable segmentation function, a (convolutional) neural network is often used as a model and parametrized using weights and biases that are optimized during the training process.
While carrying out the training with a chosen loss function and optimization algorithm, the performance of the model is evaluated on the validation set. Several stopping criteria can be used for this, for example stopping the training when the error on the validation set increases, or training for a fixed time (and recording the network that gives the best results on the validation set). To avoid any bias towards the training and validation data, the accuracy of the trained model is finally assessed using the test set. \\

Since the introduction of Fully Convolutional Networks \cite{LongShelhamer}, in which successive contracting convolutional layers are utilized for pixel-wise semantic segmentation, many convolutional neural network (CNN) architectures have been proposed that can be used for the object segmentation task. U-Net changes the FCN architecture by - along with downsampling operators and skip connections - introducing upsampling operators instead of pooling operators, giving it an U-shaped appearance \cite{RonnebergerFischer}. Similarly, Deconvnet \cite{NohHong} also introduces an auto-encoder structure with deconvolution and unpooling operations (without skip connections). The success of these methods on medical image segmentation and object detection spawned other commonly used CNN architectures for segmentation such as SegNet \cite{BadrinarayananKendall}, RefineNet \cite{LinMilan}, PSPNet \cite{ZhaoShi}, and Mask R-CNN \cite{HeGkioxari} for instance segmentation. Although some of the listed architectures need relatively few training examples for successful segmentation, the annotation of these examples still requires considerable efforts.

\subsection{Proposed workflow for training data acquisition}

\begin{figure}[H]
    \centering
	\includegraphics[width=0.99\textwidth]{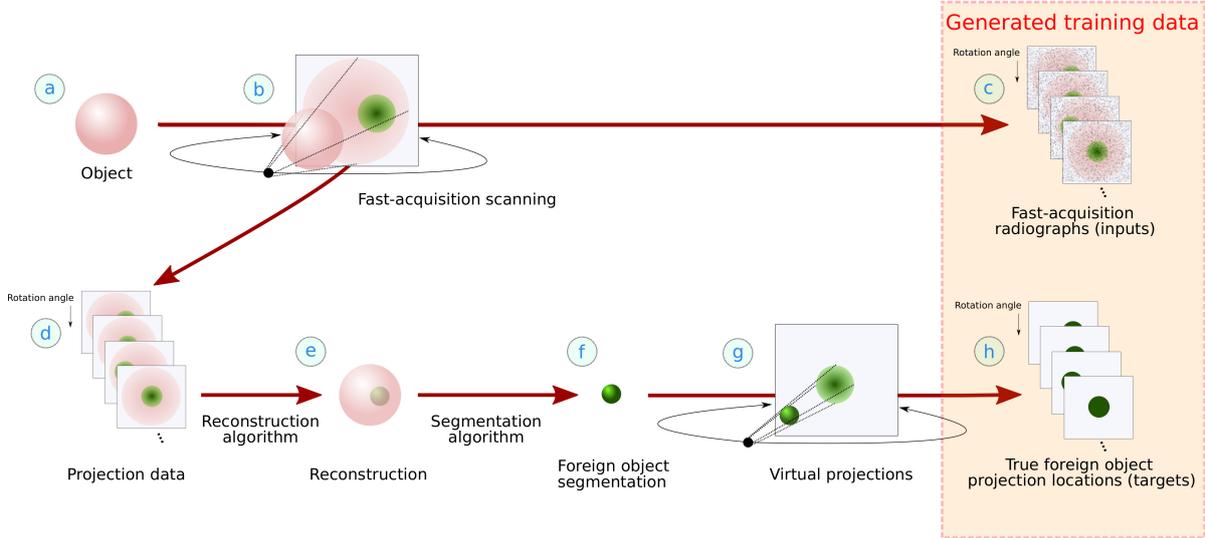}
    \caption{The complete workflow of data acquisition (\textbf{a},\textbf{b}) and the generation of training data (\textbf{c},\textbf{h}) for deep learning driven foreign object detection, through 3D reconstruction from the CT scan (\textbf{d}, \textbf{e}), segmentation (\textbf{f}), and virtual projections (\textbf{g}). The reconstruction reveals the hidden foreign objects inside the main object. Note that the projection data (\textbf{d}) is usually just the set of fast-acquisition radiographs (\textbf{d}).}\label{fig:Workflow}
\end{figure}

Our proposed workflow for using CT to obtain annotated training images is schematically displayed in Figure~\ref{fig:Workflow}. First, we select a set of representative objects as training objects (Fig.~\ref{fig:Workflow}a). For each object, a set of fast-acquisition radiographs is collected from a set of predefined angles (Fig.~\ref{fig:Workflow}b). These fast-acquisition radiographs will form the input set of the intended training dataset (Fig.~\ref{fig:Workflow}c). The total number of examples in the resulting dataset is the number of training objects multiplied by the number of selected angles. \\

The same set of radiographs is used to carry out a tomographic reconstruction of the object and acquire high-quality CT volumetric data (Fig.~\ref{fig:Workflow}d and e). The next step is to segment the reconstructed volume such that a possible foreign object is separated from the base object (Fig.~\ref{fig:Workflow}f). This segmentation step can be automated and many methods are available to implement this \cite{LenchikHeacock}. Here, we consider volumetric segmentation methods that consist of a global thresholding step. Binary segmentation by global thresholding is defined by the following function $S:\mathbb{R} \to \{0,1\}$ that acts on every voxel $z_{ijk}$ in reconstruction volume $z$:
\begin{align*}
    S(z_{ijk}) &= \begin{cases}
    			1 & z_{ijk} \geq \theta,  \\
    			0 & z_{ijk} < \theta. \\
  			\end{cases}
\end{align*}
where $\theta$ is the segmentation threshold. The more angles and other high-quality settings are used to obtain projection data, the easier it is to accurately segment the foreign object. Easier segmentation can also be accomplished by carrying out a separate high-quality scan of the same object and making a reconstruction with these high-quality radiographs. Additionally, for segmentation, prior information about the objects can be used, such as bounding boxes on the foreign object location \cite{KernMastmeyer}. Also, 3D denoising \cite{DiwakarKumar, HendriksenPelt} can be used to remove non-foreign object pixels captured by the thresholding operation. \\
    
From the constructed foreign object segmentation, virtual ground truth projections are generated by simulating projections of the foreign objects onto a virtual detector (Fig.~\ref{fig:Workflow}g). This results in the set of ground truth images, which will serve as target images in the machine learning procedure (Fig.~\ref{fig:Workflow}h). These virtual projections need to be taken under the same angles as in the fast-acquisition scan (Fig.~\ref{fig:Workflow}b). When this procedure is repeated for all objects, this results in a large dataset with annotated training examples with which supervised machine learning can be carried out (Fig.~\ref{fig:Workflow}c and f). The trained model can then be applied to similar new objects scanned in the same fast-acquisition setting, without the need for acquisition of high-quality radiographs or CT scans.

\section{Experiments and Results} \label{section:Experiments}
\begin{figure}[!h]	
    \centering
	\includegraphics[width=0.55\textwidth]{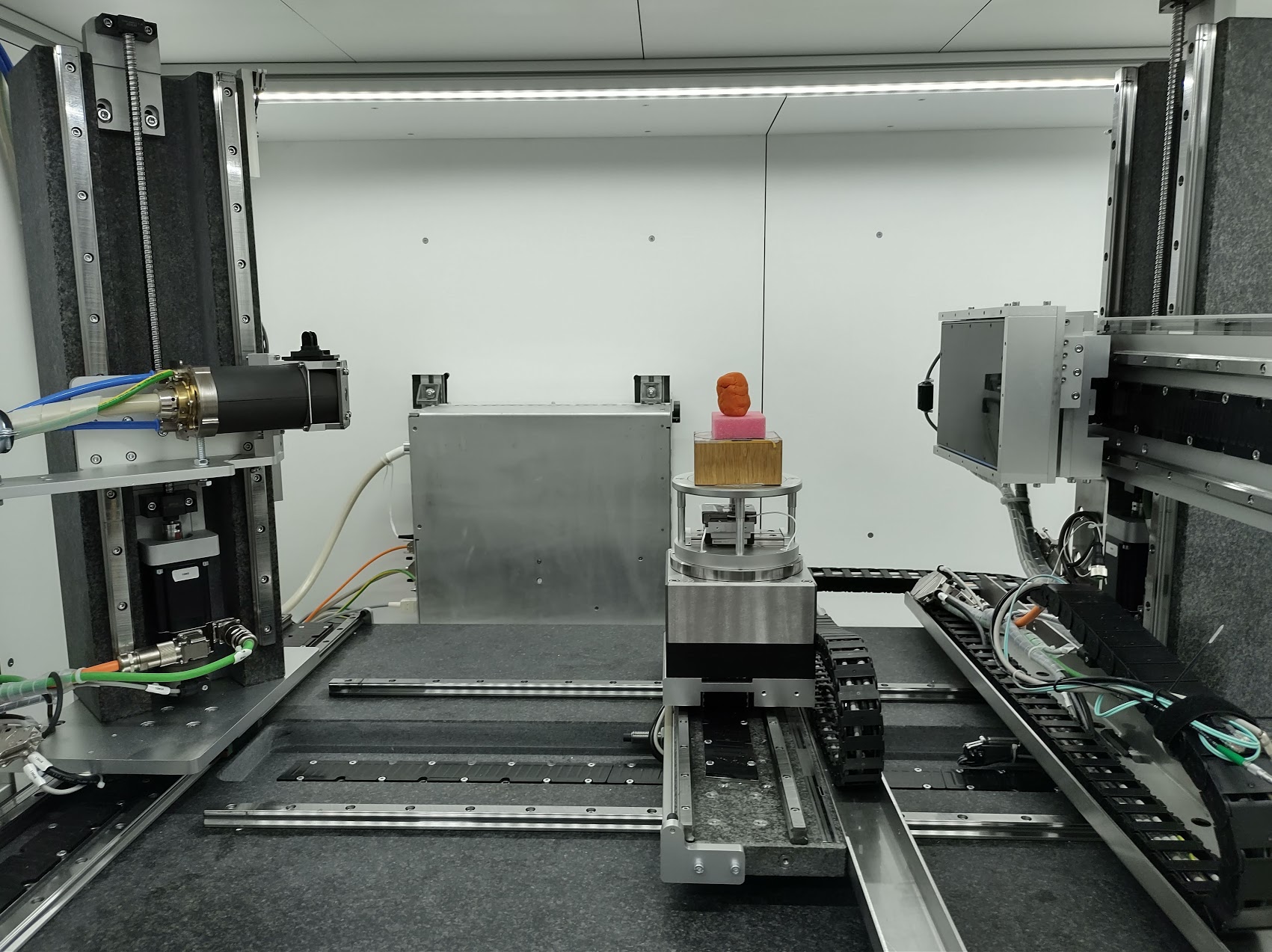}
    \caption{The scanning setup in the FleX-ray laboratory with the X-ray source on the left and the detector on the right.}\label{fig:ScanningSetup}
\end{figure}

In this section, we demonstrate the proposed workflow using the in-house FleX-ray CT system at CWI \cite{CobanLucka} (Fig.~\ref{fig:ScanningSetup}), and investigate the relation between machine learning performance and the number of training objects used. 

\subsection{Base objects and foreign objects}
\begin{figure}[H]	
    \centering
    \subcaptionbox{Play-Doh\label{fig:Play-Doh}}[0.2017\textwidth]{
		\includegraphics[width=0.2017\textwidth]{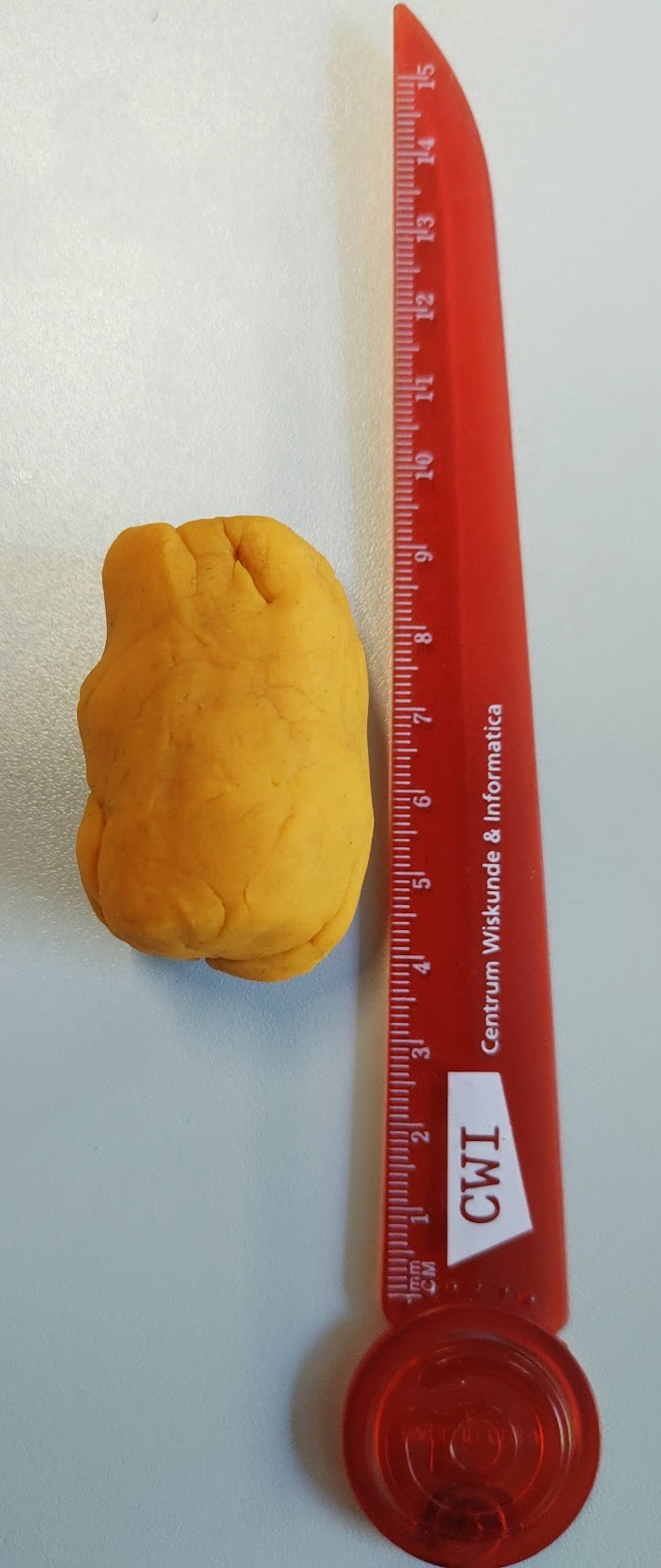}}
	\subcaptionbox{Stones\label{fig:Stones}}[0.2\textwidth]{
		\includegraphics[width=0.2\textwidth]{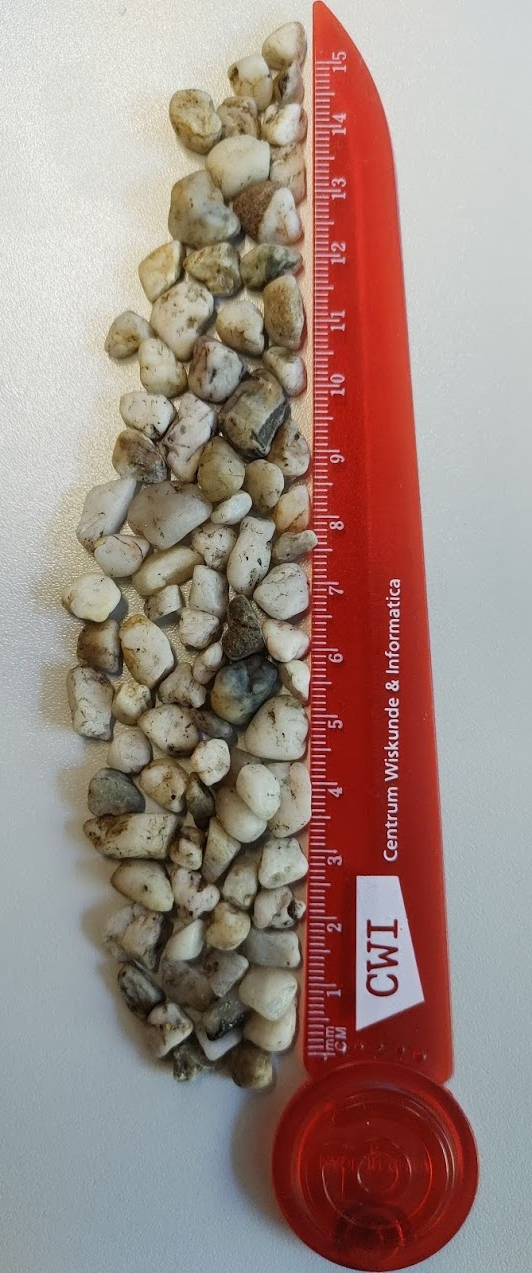}}
	\subcaptionbox{Play-Doh with stone inserted\label{fig:Play-DohStone}}[0.2\textwidth]{
		\includegraphics[width=0.2\textwidth]{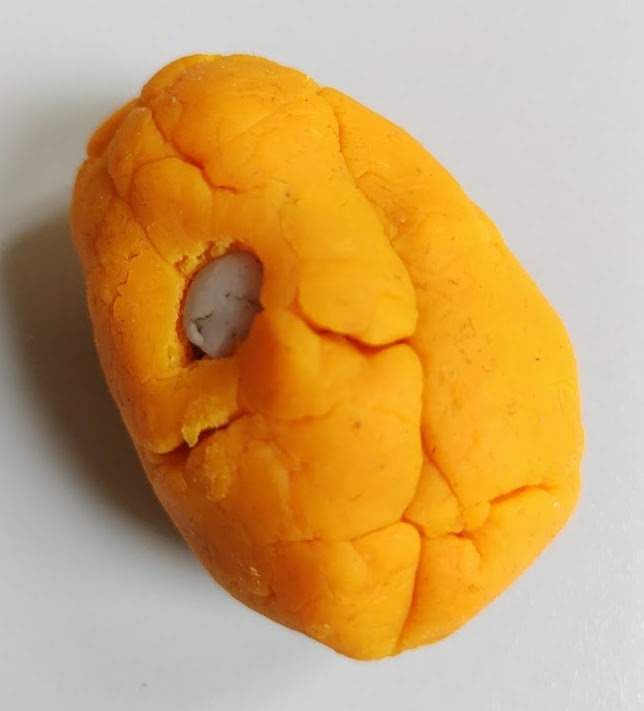}}
    \caption{An example of a base object (\textbf{a}) and examples of foreign objects (\textbf{b}) used in the laboratory experiments, as well as an example of a combined object (\textbf{c}). }\label{fig:LabObjects}
\end{figure}

As test objects, we use base objects that are created from a fixed amount of modeling clay (Play-Doh, Hasbro, RI, USA). Play-Doh is primarily made of a mixture of water, salt and flour and we therefore consider it to be a representative example of products in the food industry, where foreign objects may be pieces of stone, plastic, or metal. 
A basic shape is deformed and remolded for every object instance (Fig.~\ref{fig:Play-Doh}) in such a way that they are similar from object to object, but still exhibit some natural variation.
For the foreign objects choose to use gravel (Fig.~\ref{fig:Stones}), with the stones having an average diameter of ca. 7mm (ranging from 3mm to 11mm). These stones have slight variations in shape and material. We create $3$ objects with three inserted stones, $35$ with two stones, $62$ with one stone (Fig.~\ref{fig:Play-DohStone}) and $11$ without a stone.

\subsection{CT scanning and data preparation} \label{sec:Workflow}
\begin{figure}[H]
    \centering
	\includegraphics[width=0.3\textwidth]{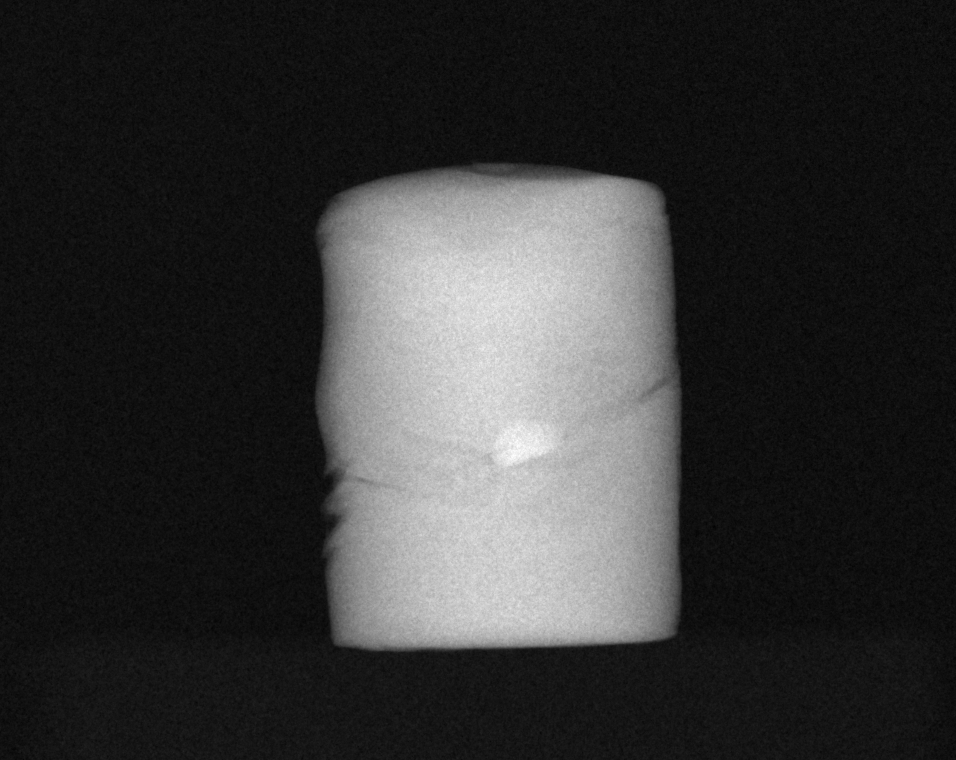} \ \
    \includegraphics[width=0.3\textwidth]{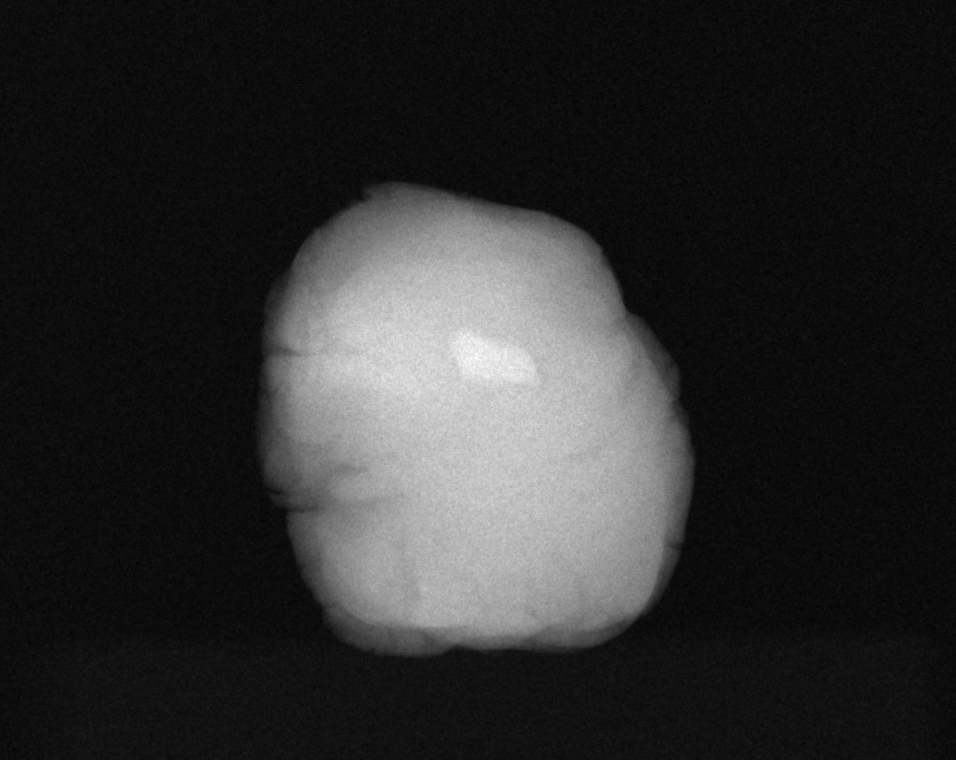} \ \
    \includegraphics[width=0.3\textwidth]{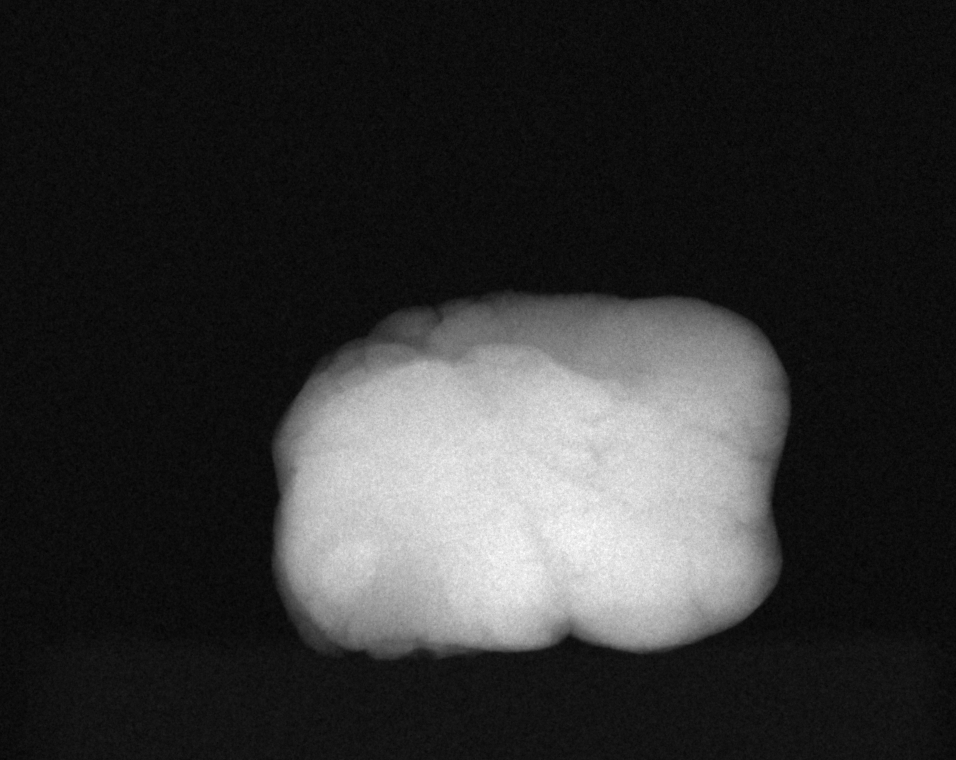} \ \
	\caption{Example of radiographs (size $965\times 760$ pixels) of three objects that are scanned. In the first and second radiographs the foreign object is clearly visible, but in the third it is more difficult to distinguish it from the base object, even though it is visible on the bottom left.}\label{fig:ProjectionDataLab}
\end{figure}

A fast CT-scan is made for each of the objects, which yields both a series of radiographs (i.e. the X-ray projections) and a reconstructed 3D volume of the object. The objects are scanned in the FleX-ray laboratory \cite{CobanLucka} (Fig.~\ref{fig:ScanningSetup}). The FleX-ray CT-scanner has a cone-beam microfocus X-ray point source with a focal spot size of 17 $\mu$m, and a Dexela1512NDT detector. The source, object and detector positions can be configured flexibly, and are arranged such that the distance between the source and detector is 69.80 cm, and the distance between the source and the object 44.14 cm. For the radiographs a voltage of 90kV with a power of 20W is used, while the exposure time is kept low at 20ms, with the intention to emulate the imaging conditions of in-line industrial systems and produce sufficiently noisy radiographs. To achieve high-quality reconstructions, 1800 projections of each object are obtained over a full $360^{\circ}$ rotation. Before and after each scan, 10 darkfield images and 10 flatfield projections are obtained. Each object is positioned in a random manner, and the cylinders may therefore be standing upright or be laying down on the long edge. Example radiographs are shown in Figure~\ref{fig:ProjectionDataLab}. Separating the projected foreign objects from the base object in these radiographs is not a trivial task, illustrating the problem of obtaining annotated training data for automated segmentation using machine learning directly from these images.

\begin{figure}[H]
    \centering
	\subcaptionbox{Reconstructed 3D volume \label{fig:ReconstructionObj3}}[0.36\textwidth]{
		\includegraphics[width=0.3\textwidth]{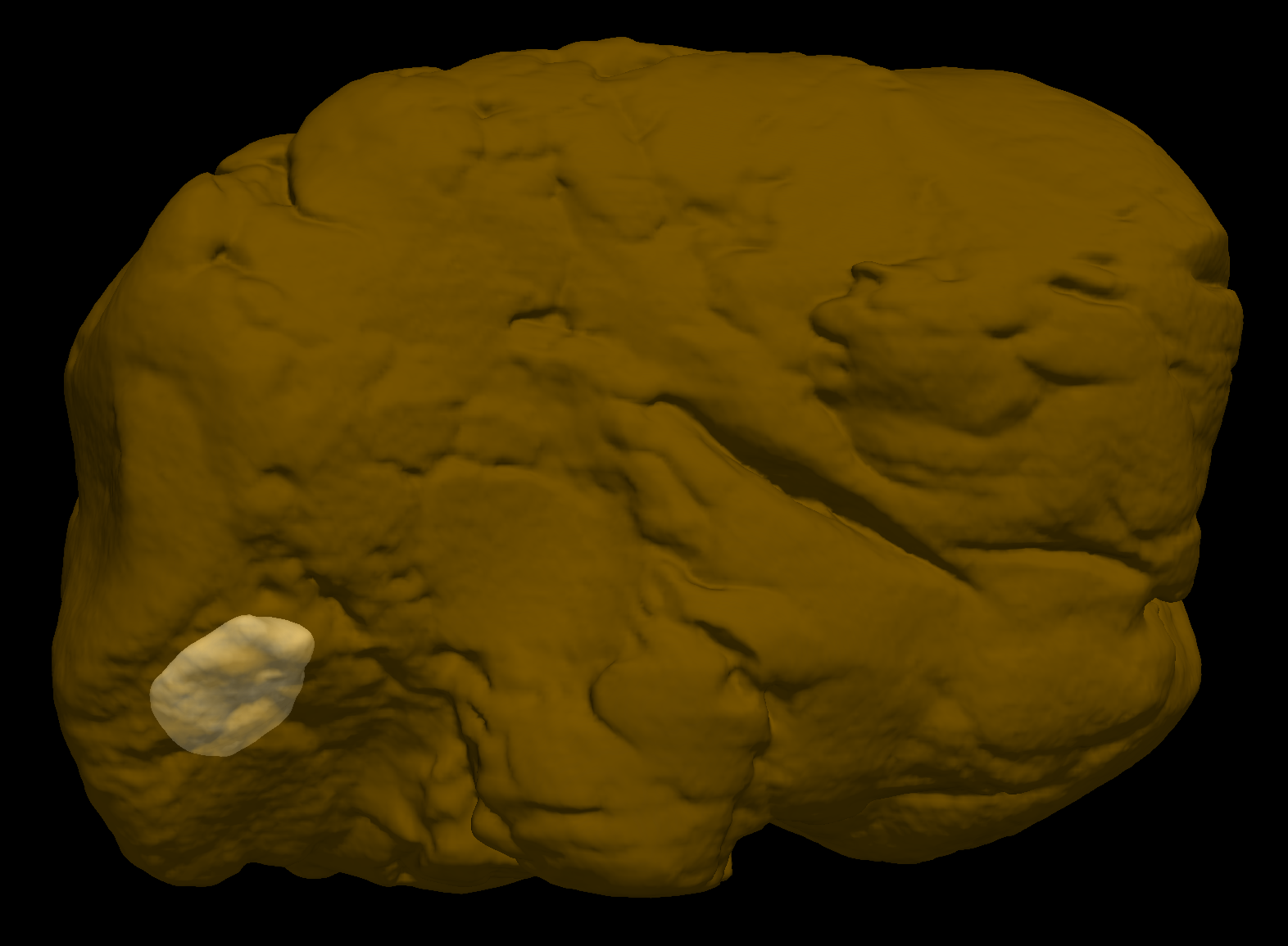}}
	\subcaptionbox{Slice of the 3D reconstruction\label{fig:ReconstructionObj3Experiments}}[0.3\textwidth]{
		\includegraphics[width=0.3\textwidth]{3DReconstructionObject3_Slice200_Crop.png}} \\
	\caption{Visualization of reconstructions by 3D rendering (\textbf{a}) and slicing (\textbf{b}) of the third object in Figure \ref{fig:ProjectionDataLab}.}\label{fig:SeparationExperiments}
\end{figure}

The Simultaneous Iterative Reconstruction Technique (SIRT) \cite{KakSlaney,SluisVorst} algorithm ($100$ iterations) as implemented in the ASTRA toolbox \cite{AarlePalenstijn2, AarlePalenstijn} is used to compute the reconstructed 3D CT volume of the object. A visualization of the reconstruction from the third object in Figure~\ref{fig:ProjectionDataLab} and its foreign object is shown in Figure~\ref{fig:SeparationExperiments}. The CT reconstruction allows to slice the object along different axes. As the CT voxel intensity is directly related to the attenuation coefficient of the material in a voxel, the segmentation task for the 3D CT volume is, in this case, much more straightforward and can be carried out by global thresholding (see Appendix for additional details on intensity value distributions). Therefore, a simple global threshold based on Otsu's method \cite{Otsu} is sufficient to segment the foreign objects.\\

From the 3D segmented objects obtained from the CT-scans, 2D segmentations for the individual radiographs are computed. This is done by computing the projections of the segmented parts with the ASTRA toolbox using the same geometric properties as used when acquiring the radiographs of the actual CT-scan. Every non-zero pixel on the detector is marked as a projected foreign object location. The result is a dataset containing $1800$ radiographs and corresponding segmented images for each object.

\subsection{Machine learning} \label{sec:MachineLearning}
We use the Mixed-Scale Dense (MSD) \cite{PeltSethian} and the common U-Net convolutional neural network architectures \cite{RonnebergerFischer} to train the task of image segmentation. For our experiments with U-Net, we have slightly changed the architecture, as we observed this improved performance in the experiments compared to the standard version. We downsample twice, with a stride of $2$. The initial number of feature maps is set to $128$, and the number of feature maps doubles for each downsampling layer. For upsampling, bilinear interpolation is used. A spatial $3\times3$ convolution operation with zero padding and a ReLU activation function are carried out before and after all~downsampling and upsampling operations. The biases and convolution weights are initialized by sampling from $\mathcal{U}(-\sqrt{k}, \sqrt{k})$, with $k = \nicefrac{1}{c_{\text{in}} \cdot a^2}$ being the range, $c_{\text{in}}$ the number of input channels and $a$ the kernel size. ADAM optimization on the average of the binary cross entropy loss and the dice loss~\cite{SudreLi, Jadon} between the data and the predictions is used for training. The network is implemented with PyTorch \cite{PaszkeGross1, PaszkeGross2}. For comparison between architectures, we also use the MSD network for training. MSD is a compact network architecture that has been demonstrated to be suitable for real-time segmentation of X-ray and CT images using relatively few training examples compared to larger networks \cite{PeltSethian}, including the U-Net architecture. We use a depth of 100 intermediate layers and width of 1 channel per intermediate layer and increase the dilation parameter repeatedly from $1$ to $10$ dilations in each layer, which are common settings for the MSD network \cite{PeltSethian, LagerwerfPelt, PeltBatenburg}. Xavier initialization is used for the convolution weights. ADAM optimization \cite{KingmaBa} is used during training on the cross-entropy loss between the ground truth and the segmented images, and the batch size of training examples is set to $10$. We use the GPU implementations in Python that are available \cite{PeltSethian, MSDGithub}. For both architectures, the learning rate is set to $0.001$ and all networks are trained on a GeForce GTX TITAN X GPU with CUDA version 10.1.243. Data augmentation is applied by rotation and flipping of the input examples. All networks are trained for 9 hours, and the network with parameters resulting in the lowest error on the validation set is used for testing. \\

With these networks, we carry out an image-to-image training from radiographs (Fig.~\ref{fig:Workflow}c) to their corresponding foreign objects segmentations (Fig.~\ref{fig:Workflow}f). For training, $60$ randomly chosen base objects containing a foreign object are used. The remaining $51$ objects are used for testing. All images are resized using cubic interpolation to $128 \times 128$ to speed up the training process (global thresholding with parameter $\theta = 0.5$ is applied to the resized ground truth images to make these binary again). We test the performance of the trained networks for different numbers of objects included in the training scheme. To compare the workflow with labour-intensive 2D data annotation, we compare the following training strategies:
\begin{itemize}
    \item \textbf{Workflow approach}: For each network, we fix the total number of training examples to $1800$. A random but fixed order of the $60$ training objects is created and the first $i$ objects among these are used for the training set. The training examples are selected from the set of radiographs and ground truths created by the workflow from these $i$ training objects in equal amounts. Every $10$th example is used for validation during training.
    \item \textbf{Manual annotation approach}: For each network, only one randomly chosen training radiograph with the corresponding ground truth is provided for each of the first $i$ included training objects. The resulting set of training examples is separated such that $\nicefrac{8}{9}$ part is used for training (rounded down to the nearest integer) and $\nicefrac{1}{9}$ part is used for validation (rounded up).
\end{itemize}

\subsection{Quality measures}
To evaluate the accuracy of the trained networks on the test set, we compute three different measures on the segmented images and the corresponding target images. The collection of these measures both assess the image segmentation accuracy and the object detection accuracy. An image segmentation accuracy is based on the classification of each pixel in the segmented image, and there are standardized ways to measure this that do not depend on any parameters \cite{GrandiniBagli}. An object detection accuracy compares connected components (groups of pixels connected by their edges) in the segmented image with the ground truth images. Although these accuracy measures require additional parameters to define the notion of detection, they are more relevant to the foreign object detection application.  \\

The first measure is an \emph{image-based average class accuracy} (also called \emph{balanced accuracy} \cite{GrandiniBagli}) to assess the accuracy of a produced segmentation. The average class accuracy of a segmented image relative to the target image is given by the sum of the true positives divided by the true positives and false negatives (the recall) of each class, averaged over the number of classes. In the binary case this becomes
\begin{align}
    \frac{1}{2}\left(\frac{\text{TP}_{\text{FO}}}{\text{TP}_{\text{FO}} + \text{FN}_{\text{FO}}} + \frac{\text{TP}_{\text{BG}}}{\text{TP}_{\text{BG}} + \text{FN}_{\text{BG}}}\right) 
    \label{eq:AvgClassAcc}.
\end{align}
Here, $\text{TP}_{\text{FO}}, \text{FN}_{\text{FO}}, \text{TP}_{\text{BG}}$ and $\text{FN}_{\text{BG}}$ are the true positive and false negative rates of the foreign object and the combined base object and background pixel classifications respectively over the entire segmented image relative to the target image. The average class accuracy as given in~\eqref{eq:AvgClassAcc} is averaged over all target images.\\

The second measure is an \emph{object based detection rate}. A \emph{connected component} is a maximal set of nonzero-valued pixels such that each pixel is reachable from another pixel in the set via a sequence of neighboring pixels in the set. Each connected component in the target image with a minimum size of $8$ pixels ($0.05\%$ of the image size) is considered as an object that should be detected. We define such an object as \emph{detected} if its pixel-wise recall relative to the segmented image is higher than a certain threshold $\eta$:
\begin{align}
    \frac{\text{TP}^{\text{tar}}_{\text{obj}}}{\text{TP}^{\text{tar}}_{\text{obj}} + \text{FN}^{\text{tar}}_{\text{obj}}} > \eta
    \label{eq:Recall}.
\end{align}
Here, $\text{TP}^{\text{tar}}_{\text{obj}}$ and $\text{TP}^{\text{tar}}_{\text{obj}}$ are the true positive and false negative pixels in the target object relative to the segmented image. In our experiments, we set $\eta = 0.3$. We define the \emph{detection rate} as the percentage of components in all target images for which condition~\eqref{eq:Recall} holds.\\

The third measure is an \emph{object based false positive detection rate}. Each connected component in the segmented image with a minimum size of $8$ pixels is considered as a potentially detected object. We define such a potentially detected object as a \emph{false positive} if its pixel-wise recall relative to the target image is lower than a certain threshold $\delta$:
\begin{align}
    \frac{\text{TP}^{\text{seg}}_{\text{obj}}}{\text{TP}^{\text{seg}}_{\text{obj}} + \text{FN}^{\text{seg}}_{\text{obj}}} < \delta
    \label{eq:RecallFP}.
\end{align}
Here, $\text{TP}^{\text{seg}}_{\text{obj}}$ and $\text{TP}^{\text{seg}}_{\text{obj}}$ are the true positive and false negative pixels in the segmented object relative to the target image. In our experiments, we set $\delta = 0.3$. We define the \emph{false positive detection rate} as the percentage of potential objects in all segmented images for which condition~\eqref{eq:RecallFP} holds. \\

\subsection{Results}

\begin{figure}[!t]
    \centering
	\subcaptionbox{Average class accuracy\label{fig:ResExp1Lab_AvgClassAcc}}[0.49\textwidth]{
		\includegraphics[width=0.49\textwidth]{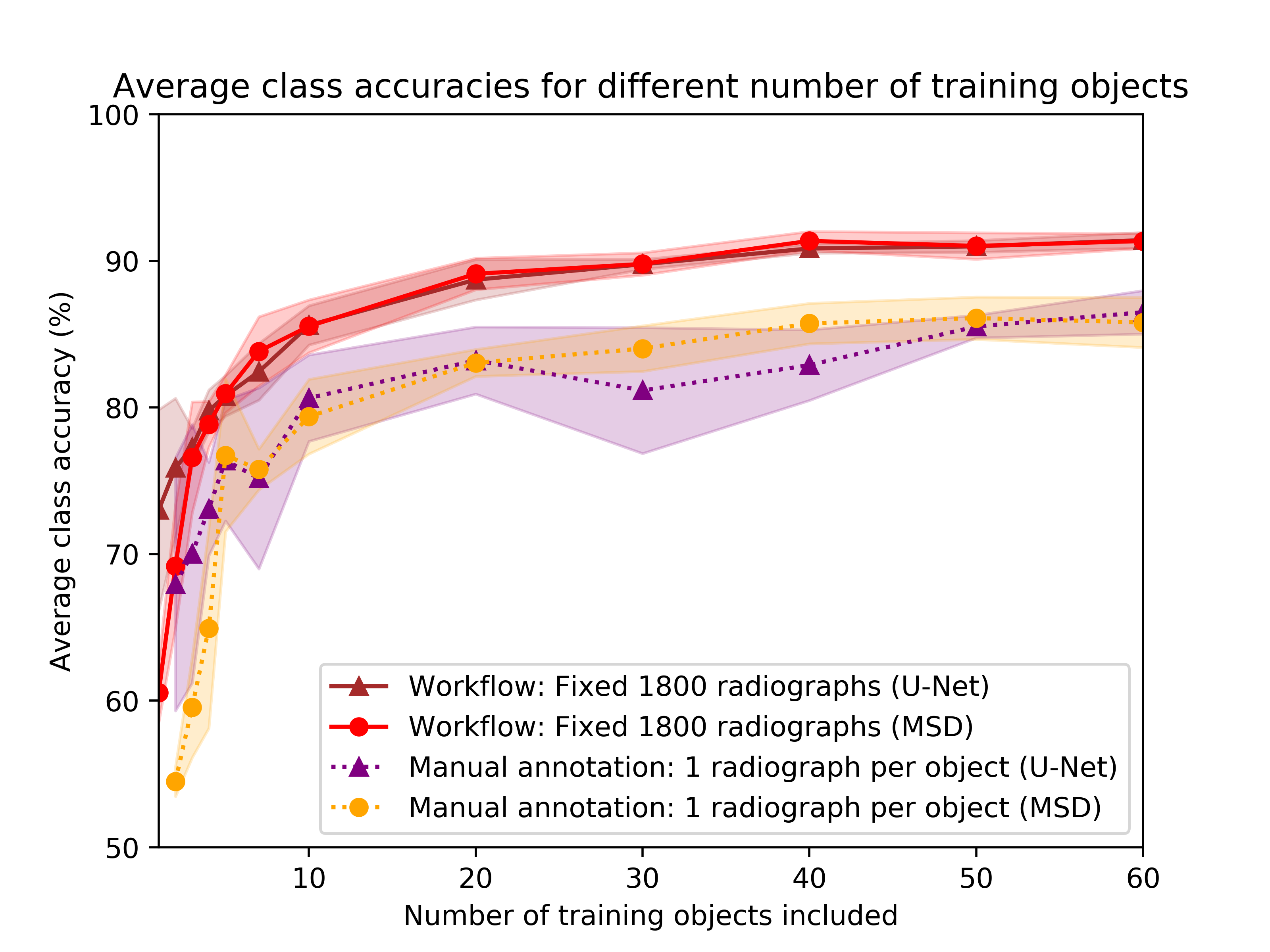}} \ \
    \subcaptionbox{Object based detection rate\label{fig:ResExp1Lab_DetAcc}}[0.49\textwidth]{
		\includegraphics[width=0.49\textwidth]{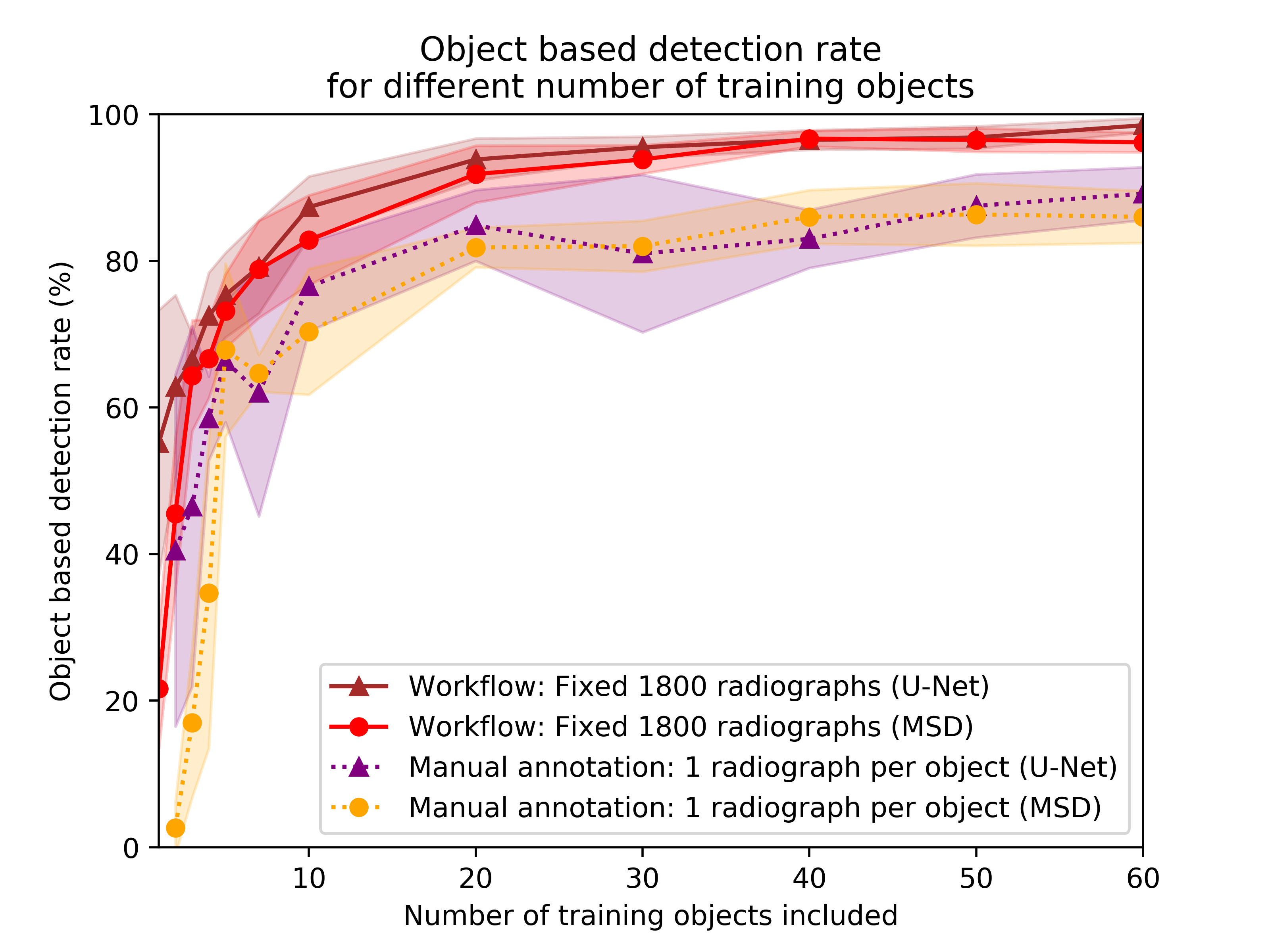}} \\
    \subcaptionbox{Object based false positive detection rate\label{fig:ResExp1Lab_FPRate}}[0.49\textwidth]{
		\includegraphics[width=0.49\textwidth]{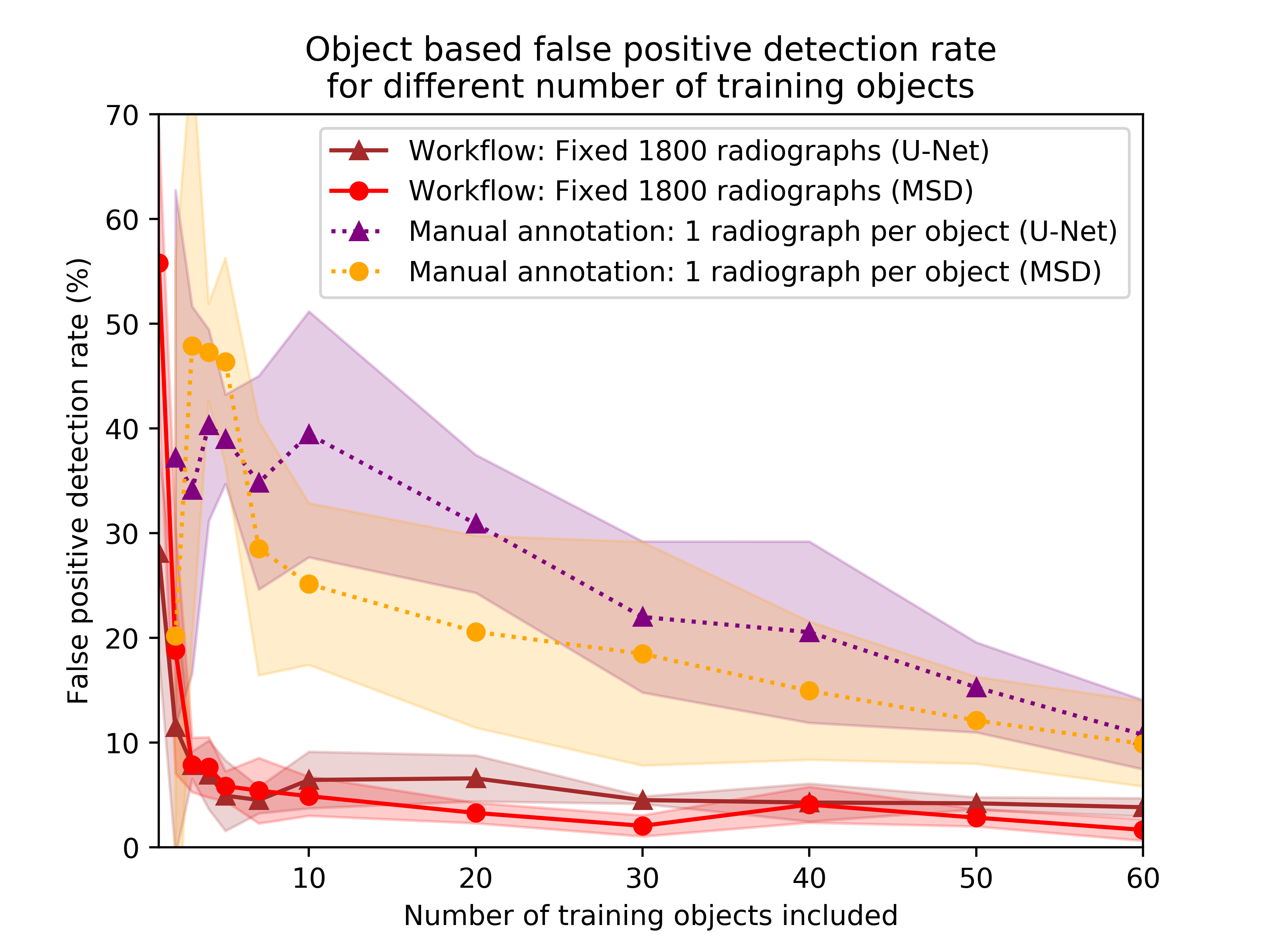}} \\[2mm]
	\caption{The average class accuracy (\textbf{a}), the object based detection rate (\textbf{b}), and the object based false positive rate (\textbf{c}) of segmentations with trained U-Net and MSD networks on laboratory data for different number of training objects. The results are shown for the fixed number of training radiograph approach and the one training radiograph per object approach. The results are averaged over $5$ trained networks, with a different training object order for each run. The shaded regions indicate the respective standard deviations.}\label{fig:ResExp1Lab}
\end{figure}

For the test set, we select a random angle and an orthogonal one for each test object, making the total number of testing radiographs $102$. We measure the average class accuracy, the object-based detection rate and the object-based false positive detection rate of segmentations created by the network on the projections from the test set. The results are given in Figure~\ref{fig:ResExp1Lab}. \\

For all measures, the quality of the foreign object segmentations on the radiographs using networks trained with the workflow data is low for a few training objects. This initially improves with the addition of relatively few training objects, but this improvement stagnates beyond 20 objects. However, the detection accuracy still shows slight improvements beyond this point, but almost completely stabilizes from $40$ objects onwards. Based on a decided accuracy goal, a certain number of objects need to be scanned and used for training to achieve that accuracy. The false positive rate decreases strongly and maintains a low level value from including $3$ objects in the training onwards. Note that the results between the U-Net and MSD architectures agree well with each other.\\

When we compare the usage of a fixed number of training radiographs among all training objects with the approach of using only one radiograph per object, we see that this leads to inferior results in all aspects. The average class accuracies and the object based detection rates are lower for all numbers of included training objects, while the false positive rates are higher. The difference between architectures only shows for the false positive detection rate, which is generally higher with the U-Net architecture. \newpage

\subsection{Laboratory experiments with many foreign objects}

\begin{figure}[H]
    \centering
    \subcaptionbox{Base object with many foreign objects\label{fig:ProjectionDataLabObjManyFO}}[0.49\textwidth]{
        \vspace{1cm}
        \includegraphics[width=0.35\textwidth]{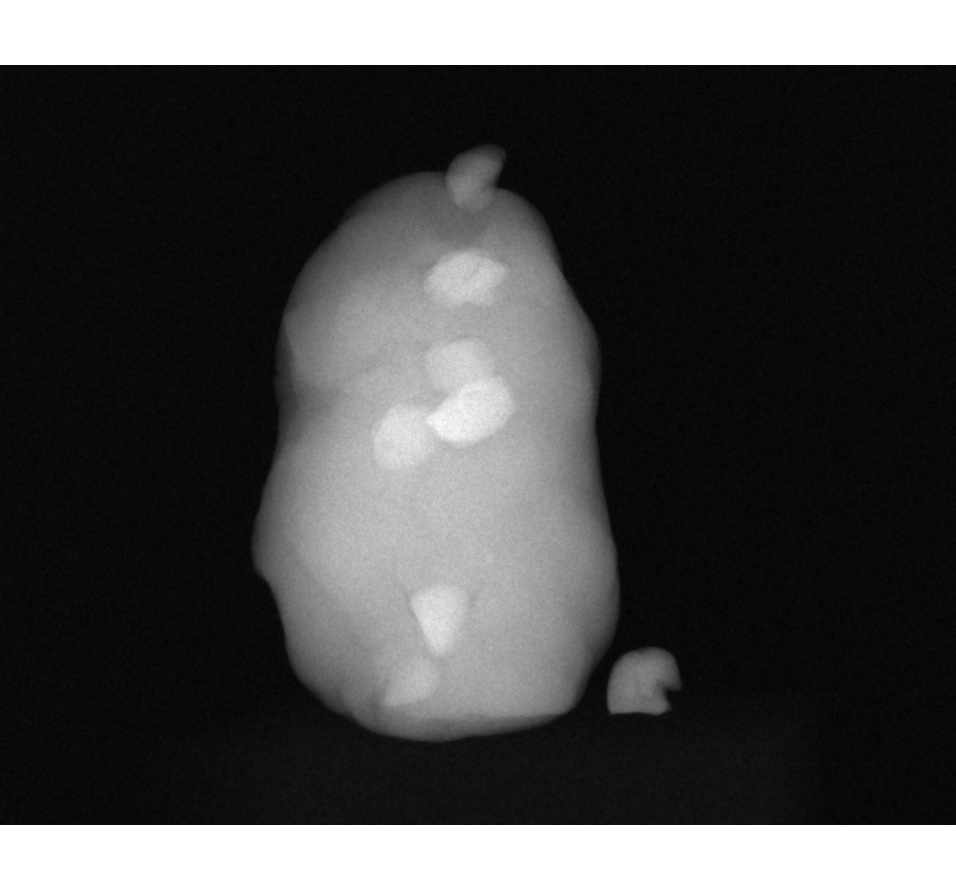}} \ \
	\subcaptionbox{Average class accuracy\label{fig:ResExpManyFOLab_AvgClassAcc_Combined}}[0.49\textwidth]{
		\includegraphics[width=0.49\textwidth]{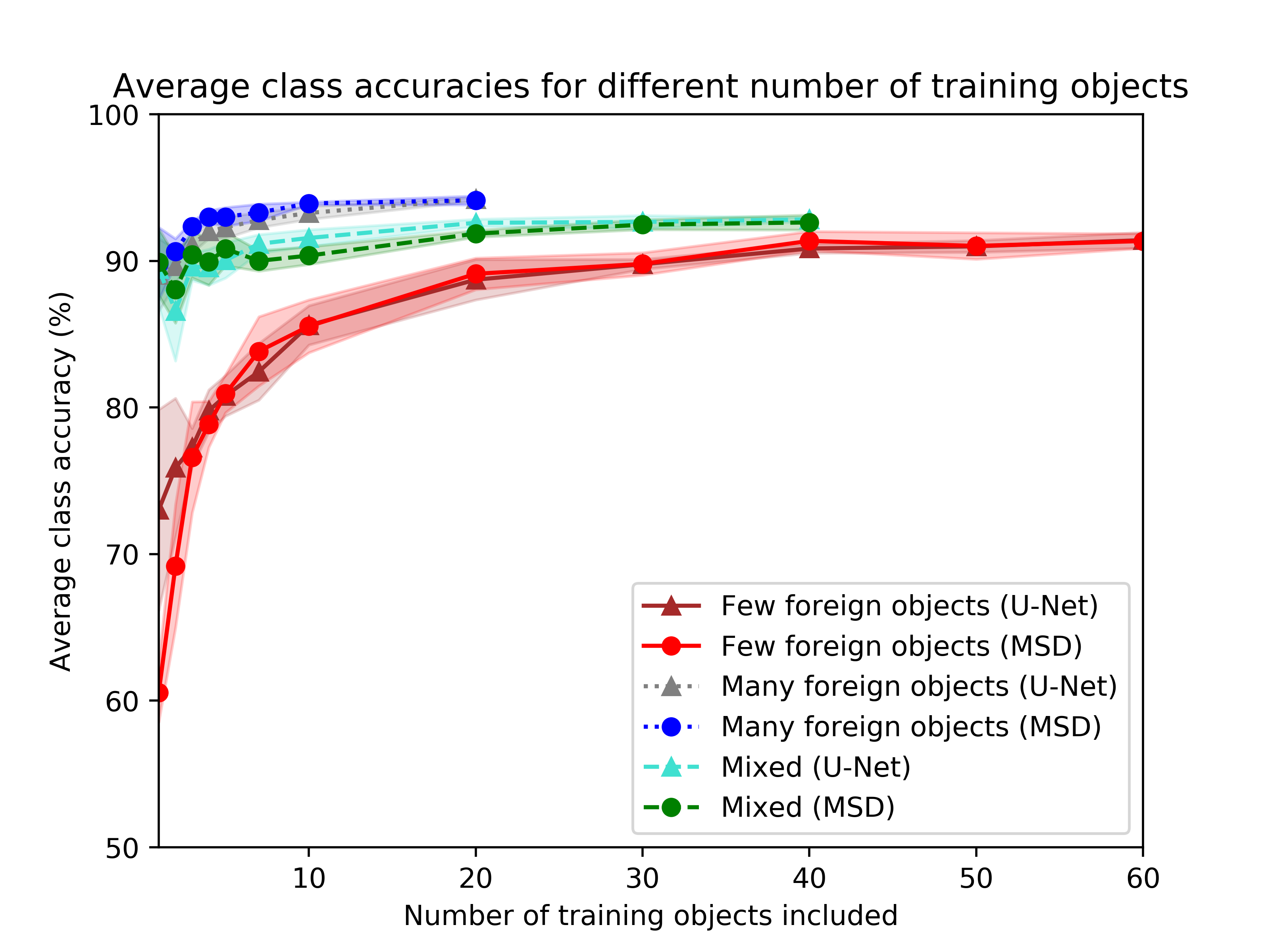}} \\
    \subcaptionbox{Object based detection rate\label{fig:ResExpManyFOLab_DetAcc_Combined}}[0.49\textwidth]{
		\includegraphics[width=0.49\textwidth]{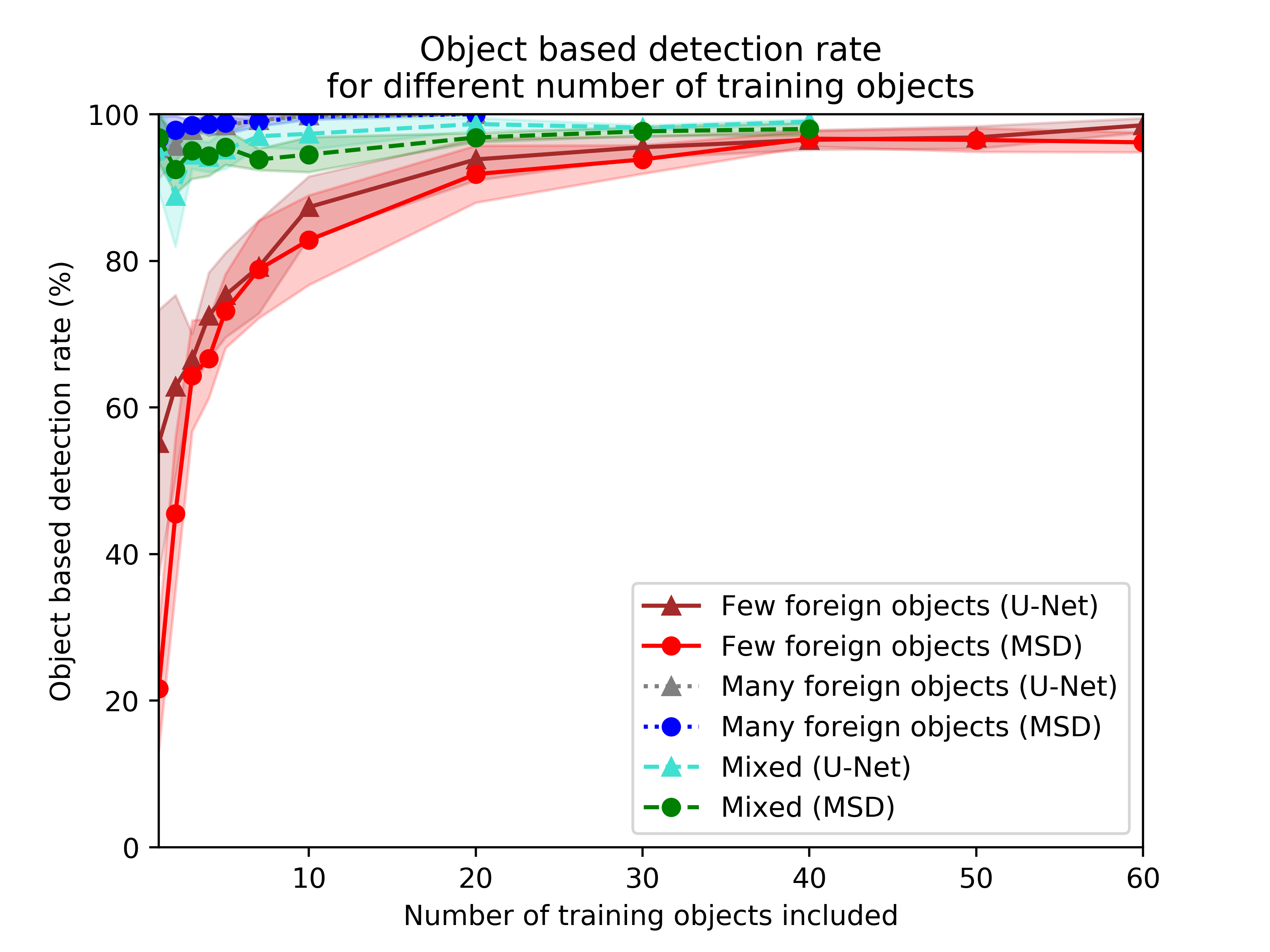}} \ \
    \subcaptionbox{Object based false positive detection rate\label{fig:ResExpManyFOLab_FPRate_Combined}}[0.49\textwidth]{
		\includegraphics[width=0.49\textwidth]{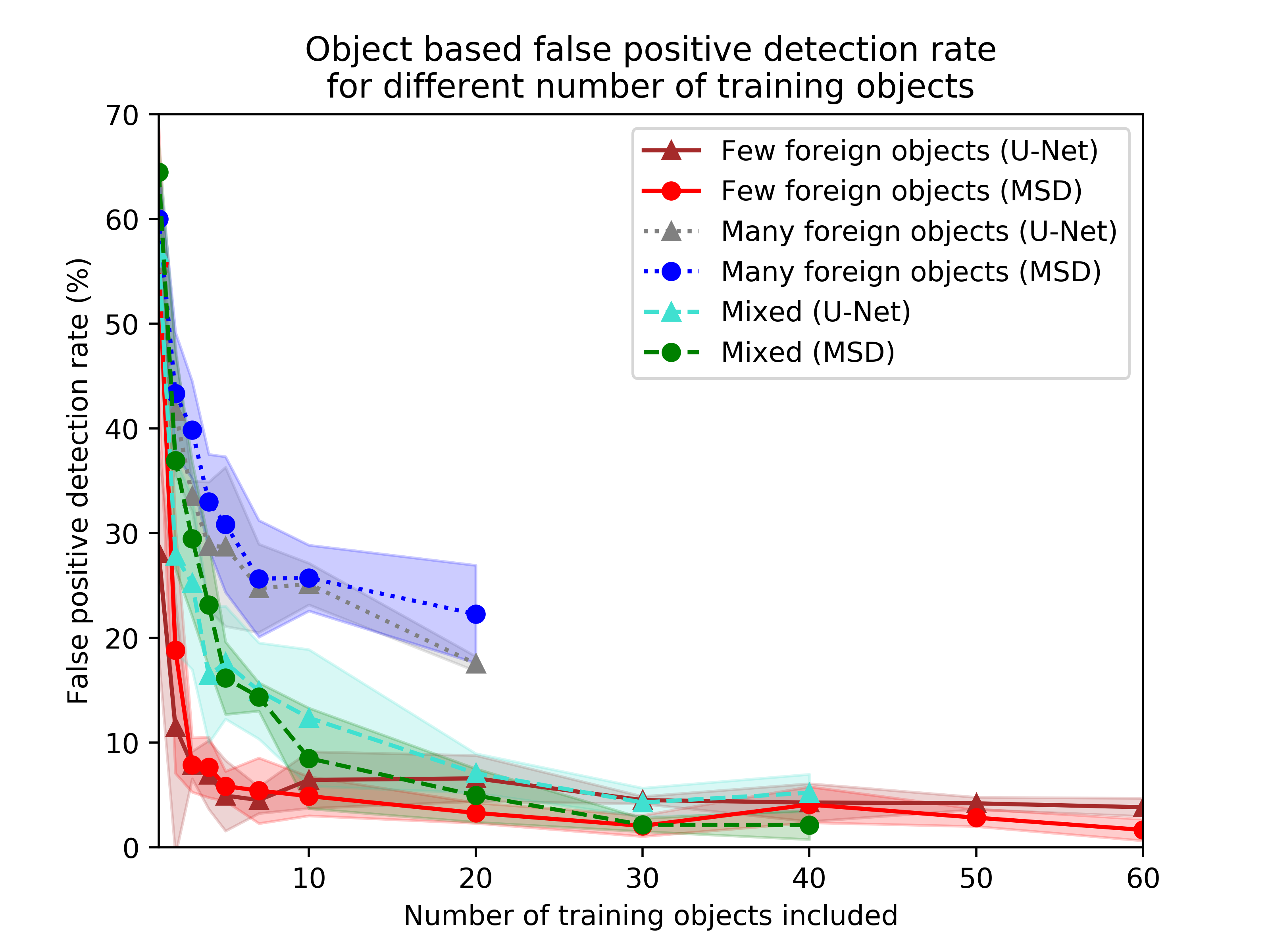}} \\[2mm]
	\caption{Example of a radiograph of a base object with many foreign objects (\textbf{a}). The eight foreign objects are vertically placed in the object, although there still may be some overlap. The average class accuracy (\textbf{b}), the object based detection rate (\textbf{c}), and the object based false positive detection rate (\textbf{d}) of segmentations with trained U-Net and MSD networks for different number of training objects and data generation strategies are shown. The results are averaged over $5$ trained networks, with a different training object order for each run. The shaded regions indicate the standard deviations.}\label{fig:ResExpManyFOLab_Combined}
\end{figure}

A natural way to reduce the number of objects used for training that need to be scanned for obtaining accurate segmentations may be to include more foreign object in the imaged objects. To test this, we repeat the experiments of the previous section, but we insert $5$ to $8$ foreign objects instead of $0$ to $2$. The foreign objects are placed within the base object such that overlapping of foreign objects in the radiographs is minimized. We have scanned an additional set of $20$ objects with these characteristics. An example of a radiograph of an object with many foreign objects is shown in Figure~\ref{fig:ProjectionDataLabObjManyFO}. We compare the following training strategies in which the workflow data comes from the following sets of training objects:
\begin{itemize}
    \item \textbf{Few foreign objects}: Base objects with $0$ to $2$ foreign objects
    \item \textbf{Many foreign objects}: Base objects with $5$ to $8$ foreign objects
    \item \textbf{Mixed}: $50\%-50\%$ mix of base objects with $0$ to $2$ foreign objects and base objects with $5$ to $8$ foreign objects.
\end{itemize}

\newpage

All networks are evaluated on the testing set from the previous section (with test objects containing few foreign objects). The average class accuracies, detection accuracies and the false positive rates of the trained neural networks with these schemes on the test set are shown in Figure~\ref{fig:ResExpManyFOLab_Combined}. From the graphs in Figures~\ref{fig:ResExpManyFOLab_AvgClassAcc_Combined} and~\ref{fig:ResExpManyFOLab_DetAcc_Combined} we see that the average class accuracies and detection accuracies are higher for the many foreign object training scheme, but Figure~\ref{fig:ResExpManyFOLab_FPRate_Combined} indicates that false positive rate is also roughly $5$ times higher. The mixed approach appears to find middle ground between the two other approaches for all measures. We see that from $20$ objects onwards the mixed approach is as good as the approach with a few foreign objects in terms of the false positive rate, while being superior in terms of average class accuracy and detection accuracy for up to $40$ training objects. This shows that including many foreign objects in the training set for detecting few to no foreign objects in the test set has limited additional value, but mixing these with examples with objects containing a few foreign objects may result in higher detection quality while maintaining a similar false positive detection rate. 

\newpage

\subsection{Robustness of the workflow}

\begin{figure}[H]
    \centering
    \subcaptionbox{Attenuation value histogram with thresholds\label{fig:Histogram_Thresholds}}[0.49\textwidth]{
		\includegraphics[width=0.49\textwidth]{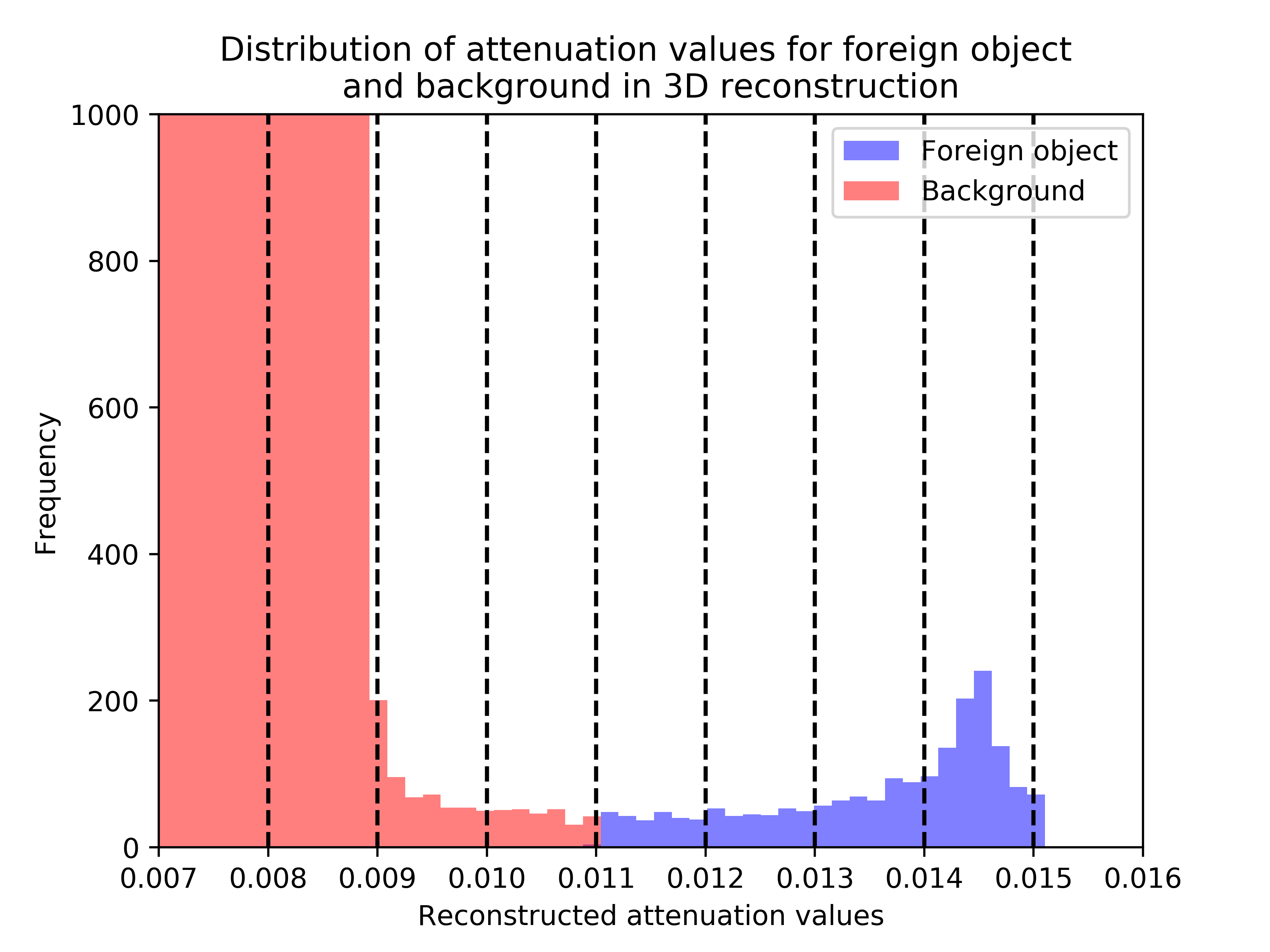}} \ \
	\subcaptionbox{Average class accuracy\label{fig:ResExpThresholdsLab_AvgClassAcc_Combined}}[0.49\textwidth]{
		\includegraphics[width=0.49\textwidth]{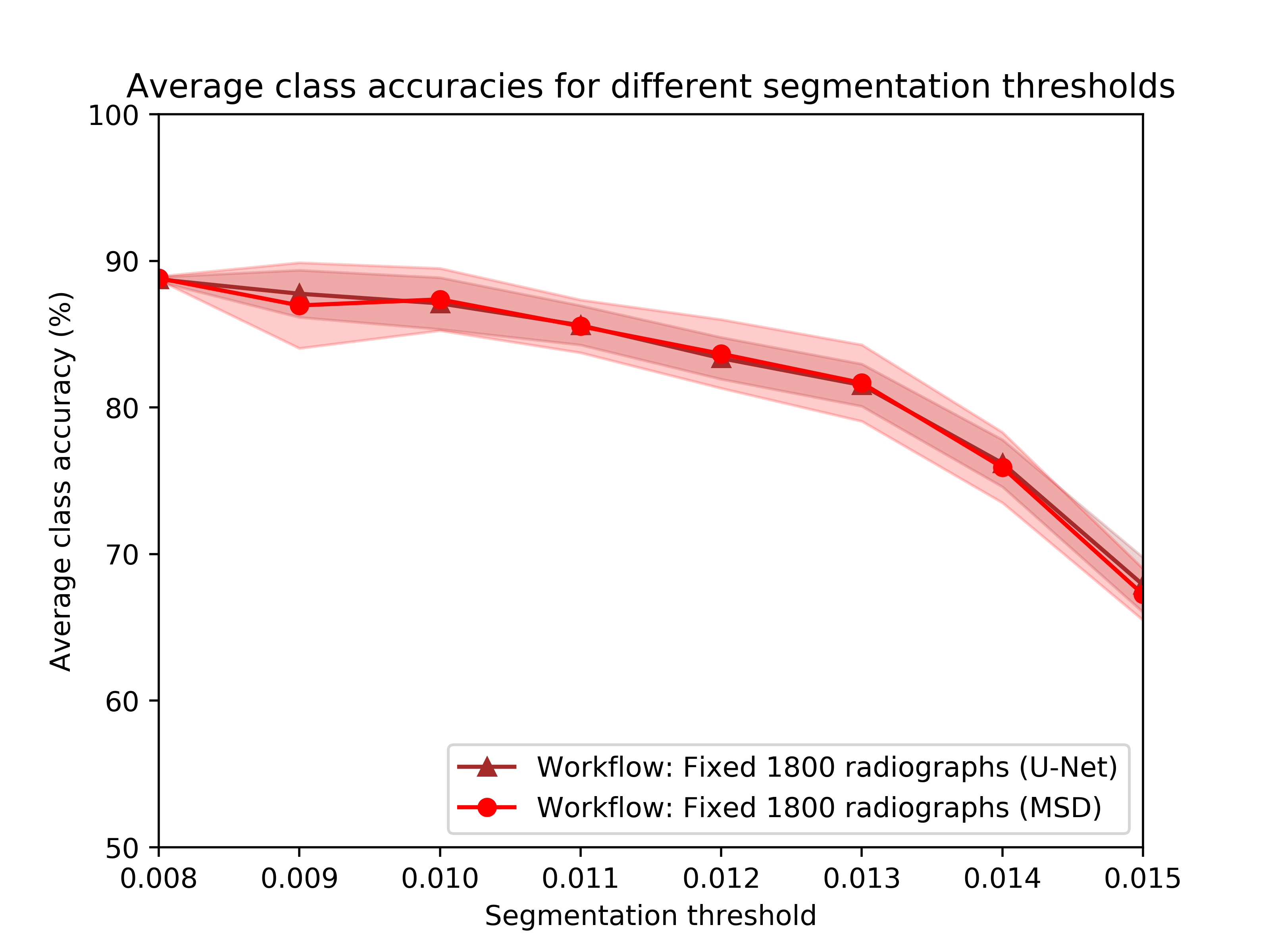}} \\
    \subcaptionbox{Object based detection rate\label{fig:ResExpThresholdsLab_DetAcc_Combined}}[0.49\textwidth]{
		\includegraphics[width=0.49\textwidth]{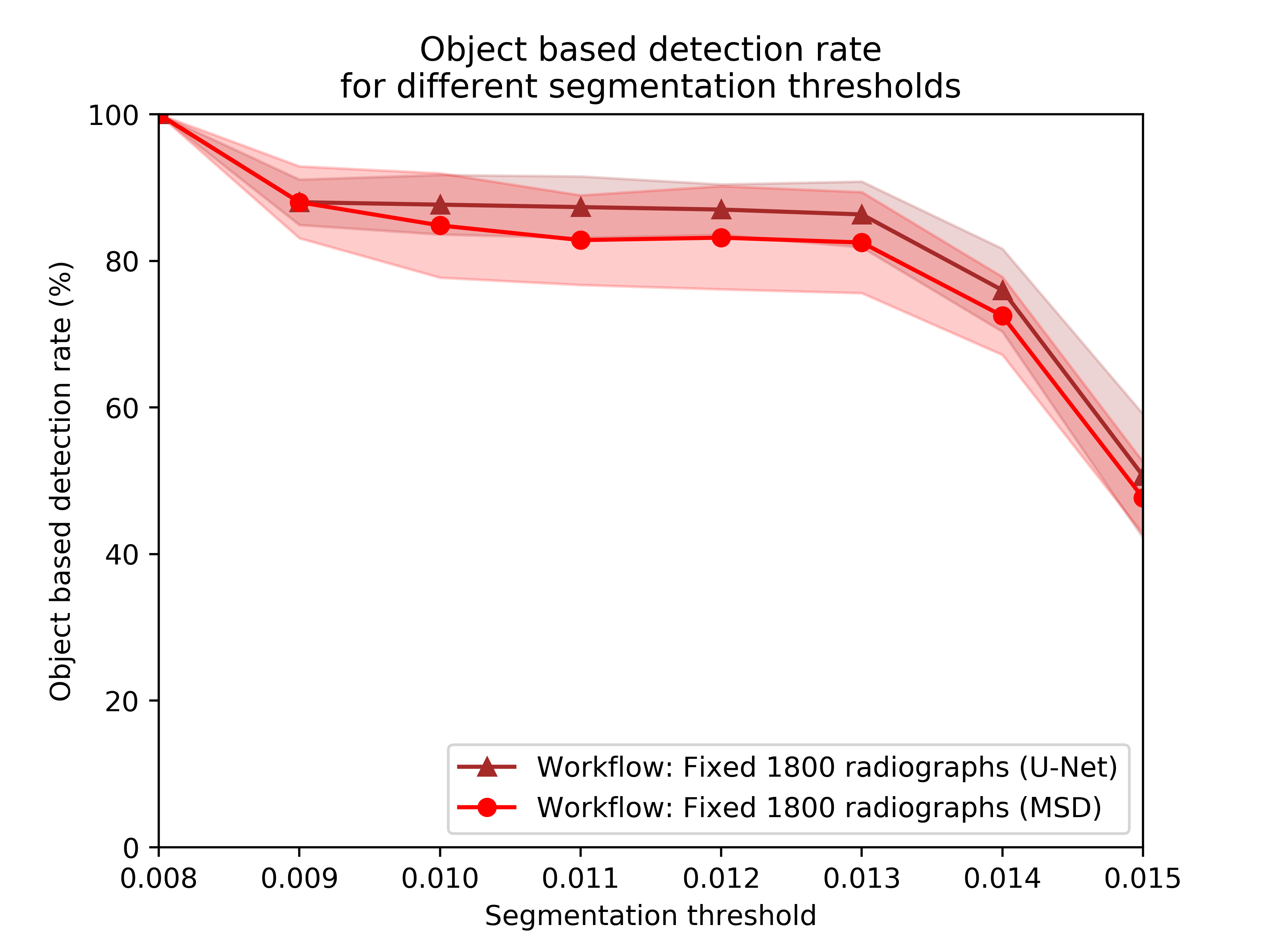}} \ \
    \subcaptionbox{Object based false positive detection rate\label{fig:ResExpThresholdsLab_FPRate_Combined}}[0.49\textwidth]{
		\includegraphics[width=0.49\textwidth]{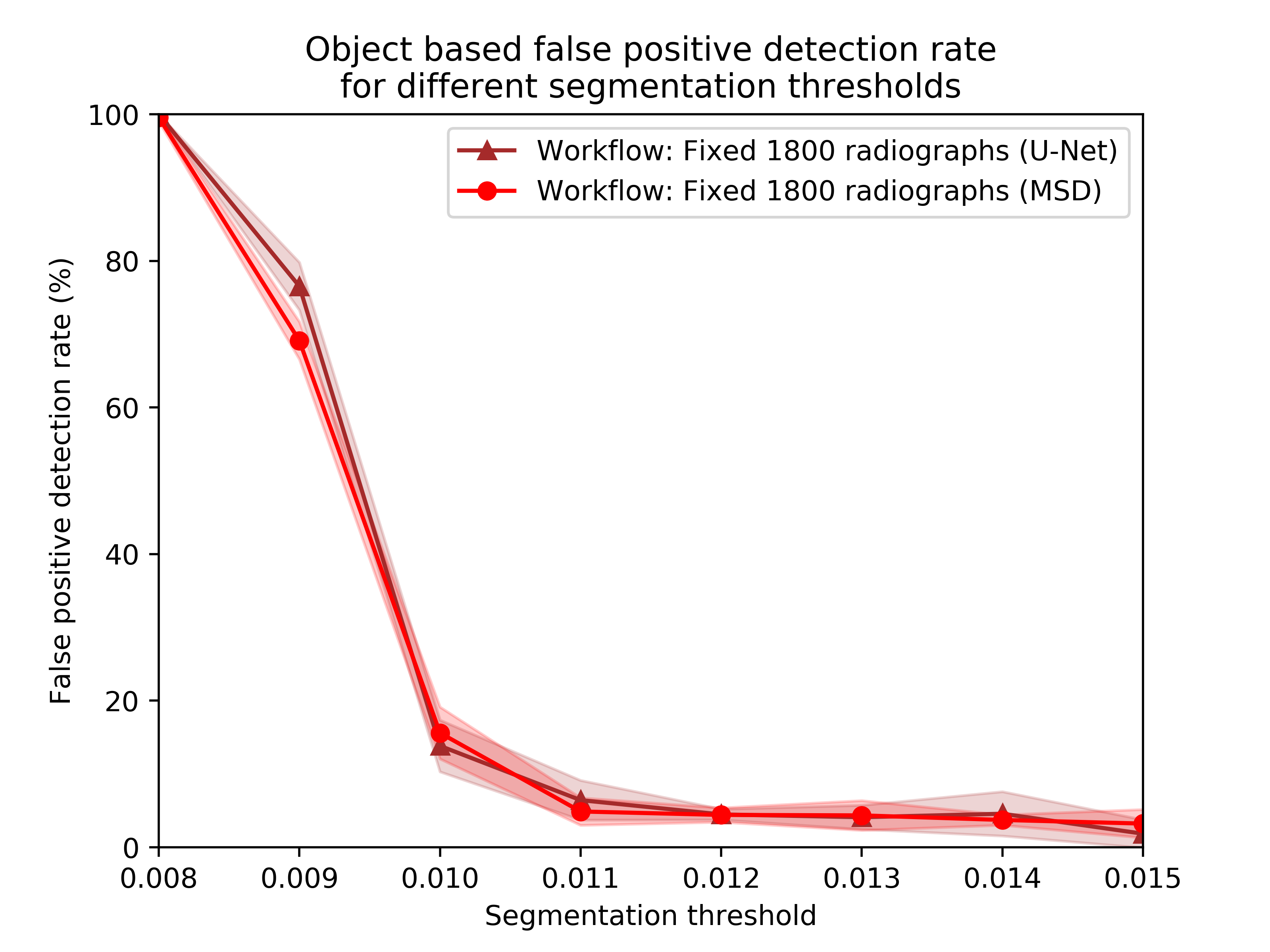}} \\
	\caption{The eight considered thresholds for the generation of the training datasets in the workflow, drawn in the histogram of attenuation values of the third object in Figure~\ref{fig:ProjectionDataLab}, and the average class accuracy (\textbf{b}), the object based detection rate (\textbf{c}), and the object based false positive detection rate (\textbf{d}) of segmentations with trained U-Net and MSD networks on data resulting from these segmentation thresholds. The results are averaged over $5$ trained networks, with a different training object order for each run. The shaded regions indicate the standard deviations.}\label{fig:ResExpThresholdsLab_Combined}
\end{figure}

In the previous experiments, the trained networks are tested on a set of projections that are generated using the same 3D segmentation threshold parameter in the workflow as in the generation of the data for the training and validation sets. To assess the robustness of the workflow to different segmentation parameters, we generate the training datasets with different values of the segmentation parameter $\theta$ (see Figure~\ref{fig:Histogram_Thresholds}). For each of these values, networks are trained and assessed on the test set from the previous sections. The number of training objects that are included in the workflow is fixed to $10$ (which has led to equivalent results in the previous experiments as with $60$ objects in the manual annotation approach). \\

In Figure~\ref{fig:ResExpThresholdsLab_Combined}, the average class accuracies, detection accuracies and the false positive rates of the trained neural networks are shown for the different thresholds. The results for U-Net and MSD are very similar. As the threshold value increases, the average class accuracy decreases, with significantly lower values for $\theta = 0.014$ and $\theta = 0.015$. The same holds for the detection rate, but it reaches a plateau between $\theta = 0.009$ and $\theta = 0.013$ where this accuracy measure gives similar values. For low values of the threshold parameter, the false positive values are high, and from $\theta = 0.011$ and higher these are low and similar to each other. Taken together, threshold parameters between $\theta = 0.011$ and $\theta = 0.013$ lead to very similar results. We conclude that for the class of objects considered in these experiments, the workflow is robust against moderate variation of the segmentation parameter and that suboptimal segmentation methods can also be used in the workflow.

\newpage

\subsection{Simulation experiments}

\begin{figure}[H]
    \centering
	\includegraphics[width=0.99\textwidth]{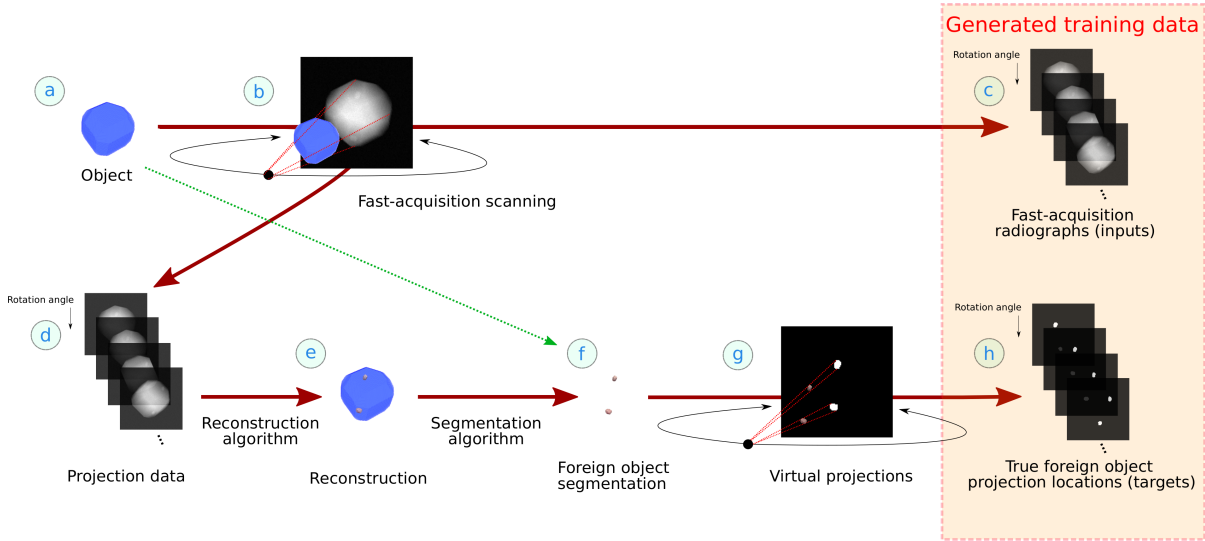}
    \caption{The complete workflow of data acquisition (\textbf{a},\textbf{b}) and the generation of training data (\textbf{c},\textbf{h}) for the simulation experiment by 3D reconstruction from the CT scan (\textbf{d}, \textbf{e}), segmentation (\textbf{f}), and virtual projections (\textbf{g}). The 3D reconstruction reveals the hidden foreign objects inside the object. The dotted green arrow (\textbf{a} to \textbf{f}) indicates that because of the simulated nature of the objects, the reconstruction and segmentation steps are skipped for the generation of ground truth for objects in the test set.}\label{fig:WorkflowApplied}
\end{figure}

In this section, we will demonstrate the workflow in a controlled simulated setting. In this way, we can verify the results with larger training and test sets when more objects are available. Furthermore, the test set previously consisted of data generated with the workflow, but in a simulated setting \lq absolute' ground truth can be created for the test set by directly projecting the simulated foreign objects (see Figure~\ref{fig:WorkflowApplied}). We verify that the proposed workflow (with CT scanning, reconstruction and segmentation) results in segmented foreign objects of which the projections are similar to absolute ground truth projections, which further supports the confidence we can have in the experimental test results. \\

\begin{figure}[H]	
    \centering
	\includegraphics[width=0.8\textwidth]{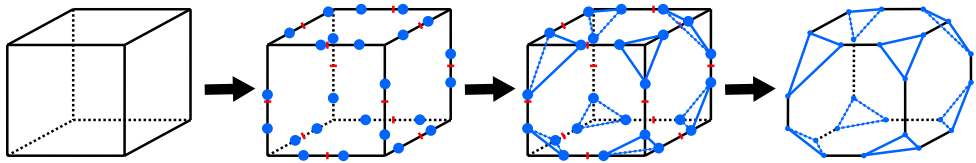}
    \caption{The process of cutting off corners with planes from cubes for creating the simulated base objects. The red stripes indicate edge midpoints and the blue dots are the randomly chosen points between those midpoints and the corners.}\label{fig:MakingCubes}
\end{figure}

We have generated a set of $500$ objects, each in an object space of $128^3$ voxels. Each object is a cube of size $64^3$ voxels, which is placed in the center of the volume. To create sufficient variety among the objects, the cube is cut off by eight planes. For each corner of the cube, a plane is created by selecting points on each of the three outgoing edges of the corner, randomly between the corner point and the midpoint of that edge. The pixels are cut off whose location is on side of the plane opposite to the center of the cube. See Figure~\ref{fig:MakingCubes} for a visualization. Additionally, we rotate the resulting object with random angles around all axes. After that we include a foreign object as an ellipsoid with a radius randomly chosen between 3 and 7 voxels at a random location within or on the edge of the base object. These ellipsoids have a random orientation as well. As a result, the foreign objects vary in shape, size, orientation and location. With $50\%$ probability, we include two of these foreign objects instead of one in the base object. \\

\begin{figure}[!t]	
    \centering
    \includegraphics[width=0.19\textwidth]{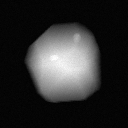}
	\includegraphics[width=0.19\textwidth]{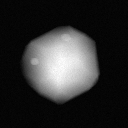}
	\includegraphics[width=0.19\textwidth]{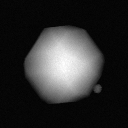}
	\includegraphics[width=0.19\textwidth]{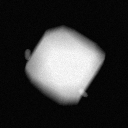}
	\includegraphics[width=0.19\textwidth]{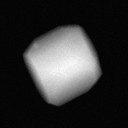}
    \caption{Example radiographs of five simulated objects. Foreign objects are located at various positions in or on the border of the base object in the radiographs, and there can be one or two of these present.}\label{fig:ProjectionExamples}
    \end{figure}

\begin{figure}[!b]
    \centering
	\subcaptionbox{Average class accuracy\label{fig:ResExpSimulated_AvgClassAcc_Combined}}[0.48\textwidth]{
		\includegraphics[width=0.48\textwidth]{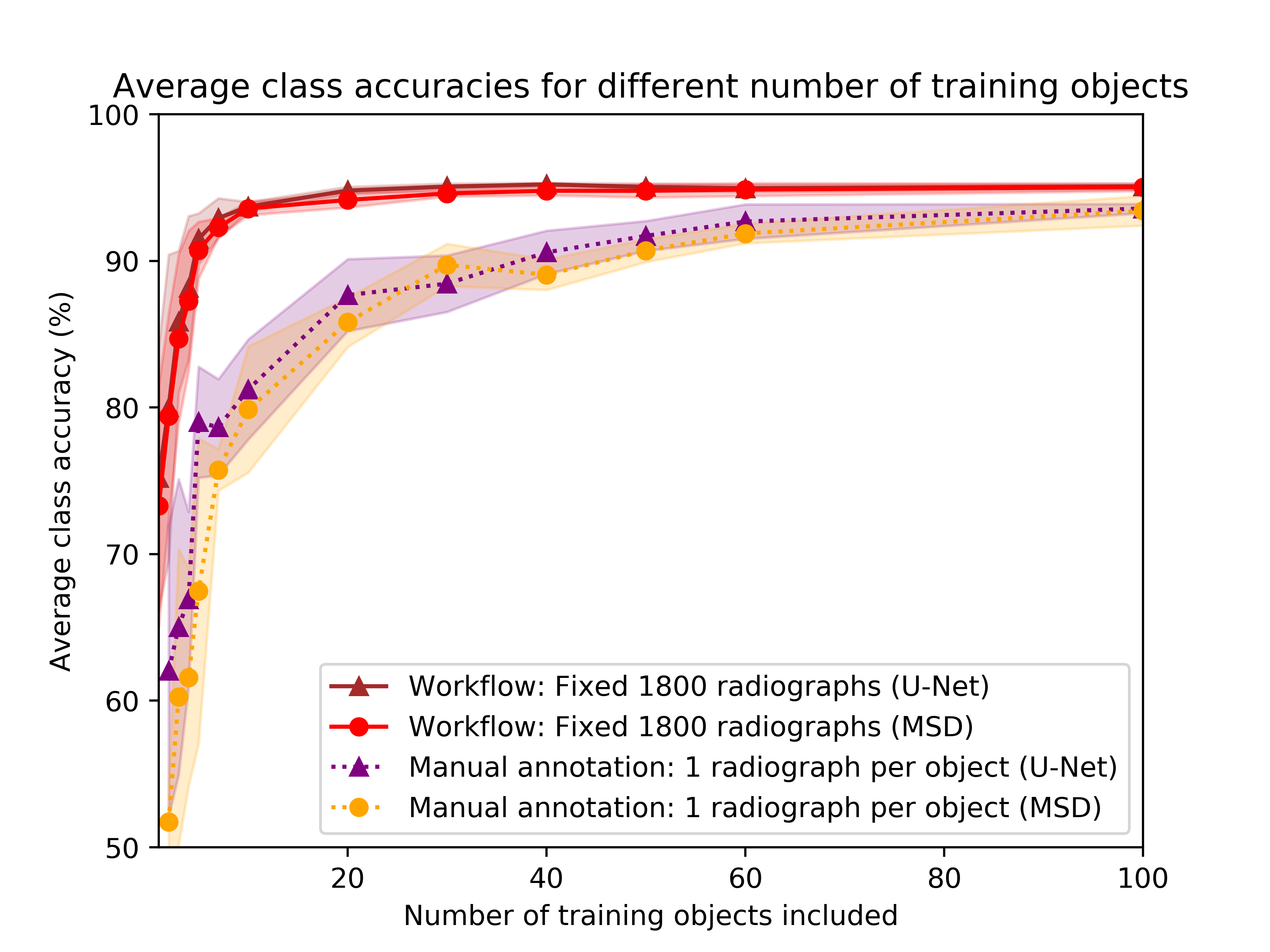}} \ \
    \subcaptionbox{Object based detection rate\label{fig:ResExpSimulated_DetAcc_Combined}}[0.48\textwidth]{
		\includegraphics[width=0.48\textwidth]{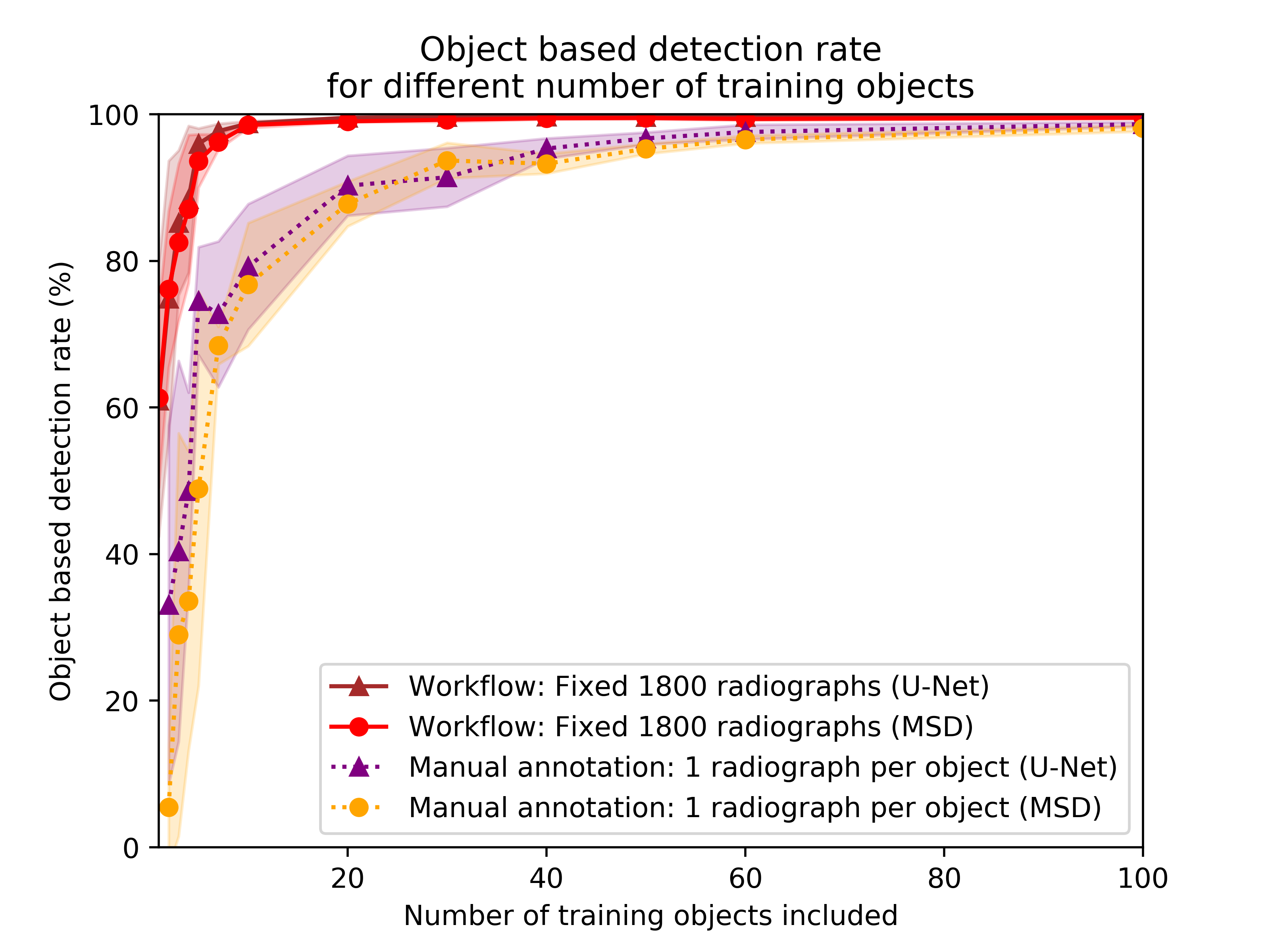}} \\
    \subcaptionbox{Object based false positive detection rate\label{fig:ResExpSimulated_FPRate_Combined}}[0.48\textwidth]{
		\includegraphics[width=0.48\textwidth]{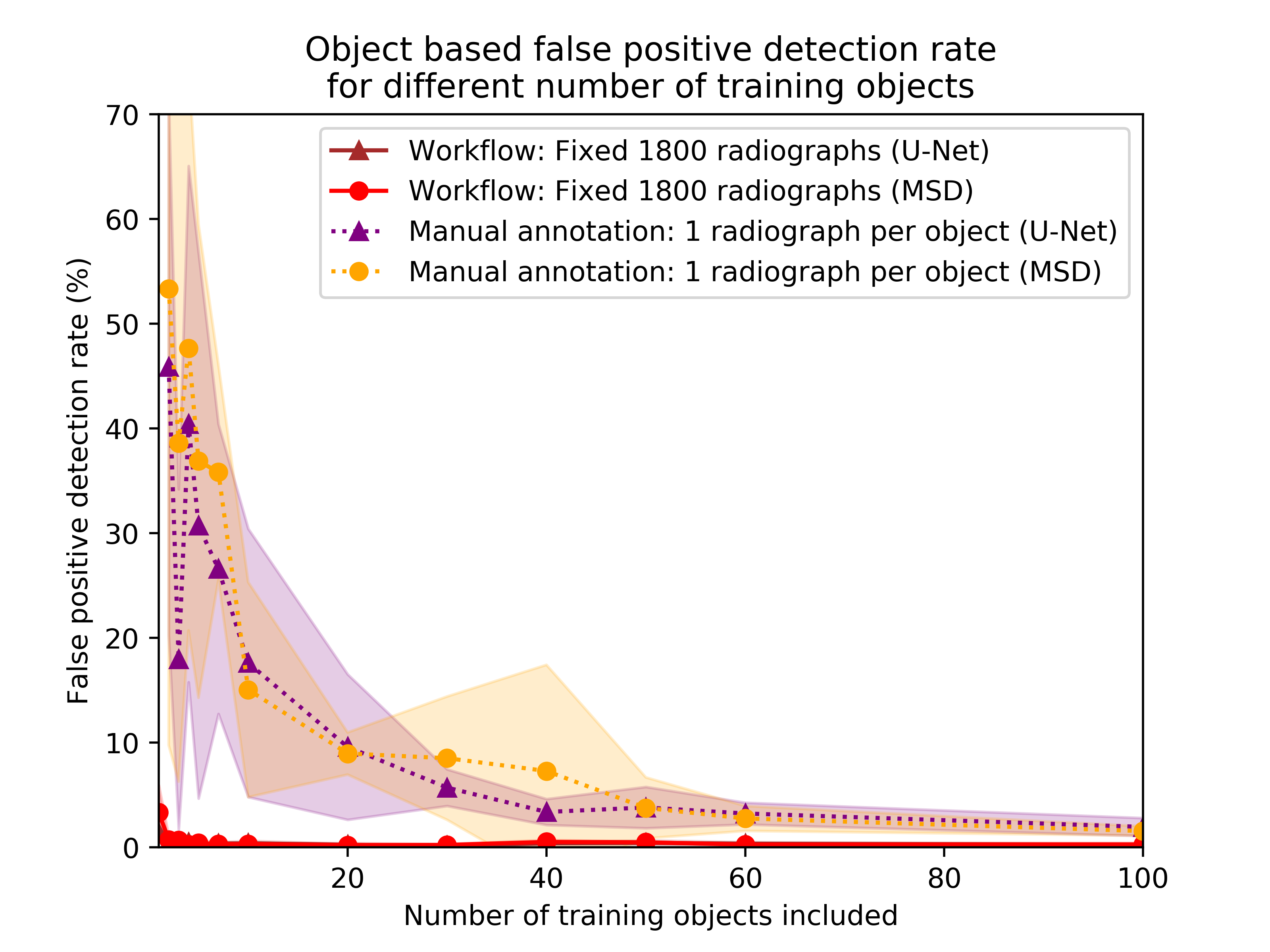}} \\[2mm]
	\caption{The average class accuracy, the object based detection rate, and the object based false positive detection rate of segmentations with trained U-Net and MSD networks on simulated data for different number of training objects and different datasets. The results are averaged over $5$ trained networks, with a different training object order for each run. The shaded regions indicate the standard deviations.}\label{fig:ResExpSimulated}
\end{figure}

Based on the spectral properties of the assigned materials, we create simulated radiographs (Fig.~\ref{fig:WorkflowApplied}b). Details of the computation can be found in the Appendix of \cite{ZeegersPelt}. First, we make projections of each material separately by computing cone beam forward projections using the ASTRA toolbox \cite{AarlePalenstijn2, AarlePalenstijn}. From this, the simulated radiographs are computed by taking the spectral properties of each material into account (taken from the National Institute for Standards and Technology (NIST) \cite{HubbellSeltzer}). We model the foreign objects as bone and the base object as tissue for each object. We take the spectral material characteristics between $15$KeV and $90$KeV into account, and use an exposure time of $0.002$ seconds for each radiograph, for which the Poisson noise that is applied is relatively high. These settings are chosen such that there is sufficient contrast in the radiographs, but not as much that it can be very easily identified with simple segmentation methods. The simulated detector size -- and therefore the projection image size -- is $128\times 128$ pixels. Examples of radiographs from five objects are given in Figure~\ref{fig:ProjectionExamples}. \\

A total of $100$ objects are reserved as training objects, while the other $400$ objects are reserved for testing. For each training object, the ground truth corresponding to each radiograph is generated with the workflow, with the same strategy and parameters as in Section~\ref{sec:Workflow}. Global thresholding with parameter value $\theta = 0.04$ is used for the reconstructions. For each test object, the `absolute' ground truth corresponding to each radiograph is generated by directly projecting the virtual foreign objects (Fig.~\ref{fig:WorkflowApplied}a and f), thereby skipping the reconstruction and segmentation steps. The projections are segmented such that every non-zero pixel on the detector is a projected foreign object location. \\

To verify that the direct use of the generated 3D volumes results in very similar ground truth projections compared to when to workflow is followed, the resulting ground truth projections are compared for the training set. The Jaccard index between the resulting ground truth pairs, averaged over all projection angles for all $100$ training objects, is $0.961$ for SIRT with $100$ iterations. This result indicates that the resulting ground truth projections resulting from both approaches are very similar. \\

To further confirm this, the training of networks as described in  Section~\ref{sec:MachineLearning} is repeated with the simulated projections, with the trained networks this time being evaluated on the test set with `absolute' ground truth. The results for the three measures are given in Figure~\ref{fig:ResExpSimulated}, and are in accordance with the experiments with the laboratory data. A notable difference is that the average class accuracy and detection accuracy reach their maximum values for a relatively lower number of training objects (and the same goes for the minimum value of the false positive rate). This is most likely because the simulated objects are less complex, resulting in radiographs with less complicated structures. Nevertheless, the results again show inferior results for the approach where one radiograph per training object is used, since $100$ objects are needed to reach similar quality measure values as for the workflow with only $7$ objects.

\section{Discussion} \label{section:Discussion}

Overall, the graphs presenting the foreign object detection accuracies in Section \ref{section:Experiments} indicate an increase of segmentation and detection accuracy with increasing the number of objects from which the training data is created. The accuracies initially increase strongly with the number of training objects but this increase decays when the number of training objects is further increased. The maximum detection accuracy that can be achieved depends on the nature of the foreign detection problem. For instance, if the X-ray flux is low and the noise is high, foreign objects are more difficult to detect from the radiographs. In the case of the laboratory experiments, foreign objects are difficult to detect when the cylindrical shape was located with the long edge on the ground and oriented orthogonal to the detector. The radiographs should contain sufficient discriminatory information such that foreign object detection with deep learning is possible. Additionally, for the dataset to be suitable for supervised machine learning, the ground truth should also be of sufficient quality, although this seemed to be less of an issue in our experiments as we observed no negative effects from occasional noise in the ground truth on the training and detection accuracy.\\

With the above considerations in mind, the workflow is designed to be modular. Every stage of the proposed workflow can be designed according to the available data-acquisition equipment, the intended detection accuracy, the type of base objects and foreign objects, and the available computer memory, among other things. We highlight some possible considerations for every stage:
\begin{itemize}
    \item Objects (Fig.~\ref{fig:Workflow}a): The set of objects can be enlarged or diversified when the accuracy of the trained neural network is not satisfactory. Also, more objects can be added to obtain a more diverse representation of objects when a more diverse array of objects or orientations are considered to be subjected to X-rays in the industrial application, such as on a conveyor belt. When a completely new type of objects is considered, these objects should be added to the workflow as well.
    \item Scanning routine (Fig.~\ref{fig:Workflow}b): In our experimental setting we have used data resulting from low exposure times as input for both the neural networks and the reconstruction algorithm. If the foreign objects turn out to be too difficult to separate in the reconstructions, more scanning angles may be considered. Additionally, if the factory settings are allowed to be altered, higher fluxes, different tube voltages or longer exposure times can be used to obtain radiographs of higher quality, as long as the processing times remain acceptable. Also, more discrimination can be achieved by applying spectral imaging (dual-energy \cite{RebuffelDinten} or multi-energy imaging \cite{EinarssonJensen, SiBar, TaguchiBlevis}) such that the neural network can distinguish the foreign objects from the base objects. If changing the quality of the radiographs is not possible, a separate high-quality scan of the same object can be made under the same angles, to achieve more contrast of the foreign object in the reconstructions. The scanning routine can be carried out in any lab, as long as it done under similar conditions as in the intended industrial X-ray imaging setting. 
    \item Reconstruction algorithm (Fig.~\ref{fig:Workflow}e): Depending on the type of data, different reconstruction algorithms may be considered \cite{Buzug, HansenJorgensen}. In this work, we have used the SIRT algorithm to account for the noise in the data, but other reconstruction algorithms such as Feldkamp-Davis-Kress (FDK) algorithm \cite{FeldkampDavis} or the Conjugate Gradient method for Least Squares (CGLS) \cite{HestenesStiefel} can be considered as well. Also, when dealing with spectral or generic multi-channel data, multi-channel reconstruction methods \cite{KazantsevJorgensen, RigieRiviere, SawatzkyXu, SemerciHao, ZeegersLucka} can be used to increase the reconstruction accuracy even further. When dealing with objects that may change in time, dynamic reconstruction methods can be considered \cite{DjurabekovaGoldberg, NikitinCarlsson, HauptmannOktem}.
    \item Segmentation algorithm (Fig.~\ref{fig:Workflow}f): In this work we have used a simple global thresholding scheme, but many more segmentation methods are available, as well as approaches to reduce possible noise \cite{DiwakarKumar}, or bounding boxes when the location of the foreign object is known \cite{KernMastmeyer}. In case of multi-channel data, a multi-dimensional thresholding scheme can be used, as well as clustering methods. Discrete reconstructions algorithms that combine reconstruction and segmentation are also available \cite{BatenburgSijbers, HermanKuba}.
    \item Virtual projection (Fig.~\ref{fig:Workflow}g): When creating the virtual projection, post-processing on the generated ground truth projections can be applied to increase the training target quality, for instance by denoising the obtained ground truth projections.
    \item Supervised learning (Fig.~\ref{fig:Workflow}c and h): To validate the workflow, we have used the U-Net architecture with ADAM optimization on cross entropy loss and dice loss, as well as the MSD network with ADAM optimization \cite{KingmaBa} on the cross-entropy loss. Other neural network architectures (see Section~\ref{sec:SupervisedLearning}) can also be considered, as well as different optimization strategies and loss functions. Note that the foreign object detection problem considered in this work may be ambiguous, since for a base object containing a foreign object another base object can theoretically be constructed (without foreign object) that results the same radiograph. This constructed base object may have an unnatural shape when compared with other base objects, but if it happens, it may lead to inconsistent training data for the network. However, this possible problem is independent of the workflow and can be resolved by multi-spectral imaging or multi-angle imaging, and training the networks with multiple images from the same object resulting from these imaging methods. However, creating reconstructions with data from these advanced imaging methods would not be necessary.
\end{itemize}

The training data acquisition workflow proposed in this paper holds a possible advantage over annotation of 2D radiographs, even when perfect annotations are created. According to the results in Figures~\ref{fig:ResExp1Lab} and~\ref{fig:ResExpSimulated}, segmentation and detection accuracy can be improved by using multiple annotated radiographs for each training object. As opposed to manual annotation, with the proposed workflow many additional radiographs are obtained for each training object.

\section{Conclusions} \label{section:Conclusions}

In this research, a new workflow is proposed for generating training data for supervised deep learning for foreign object detection in an industrial setting. In this workflow, a number of representative objects are scanned using X-ray imaging, reconstructed using computed tomography, segmented and virtually projected in an objective and reproducible manner to obtain the true foreign object locations in a large set of radiographs, after which supervised machine learning can be applied to detect foreign objects with high accuracy depending on the number representative objects included. We demonstrate this workflow on both laboratory and simulated data using using neural networks for the deep learning task. Through laboratory experiments, we have verified that the workflow produces adequate target images. The introduced measures assess the quality of foreign object detection with networks trained using datasets generated with this workflow. All experiments show a consistent result in which the accuracy increases significantly with a few number of training objects, and less significantly for every additional training object. In the laboratory experiment, we consistently obtain high accuracies for detecting gravel in modeling clay with low exposure times using this workflow, demonstrating its application potential in an industrial setting.

\section*{Author contributions}

\textbf{Math\'e T. Zeegers}: Conceptualization, Methodology, Software, Validation, Formal Analysis, Investigation, Resources, Data Curation, Writing - Original Draft, Writing - Review \& Editing, Visualization. \textbf{Tristan van Leeuwen}: Conceptualization, Writing - Review \& Editing, Supervision, Project administration. \textbf{Dani\"el M. Pelt}: Conceptualization, Methodology, Software, Writing - Review \& Editing, Project Administration. \textbf{Sophia Bethany Coban}: Conceptualization, Writing - Review \& Editing. \textbf{Robert van Liere}: Conceptualization, Writing - Review \& Editing, Funding Acquisition. \textbf{Kees Joost Batenburg}: Conceptualization, Methodology, Writing - Review \& Editing, Supervision, Project Administration, Funding Acquisition.

\section*{Conflict of interest}
The authors declare no conflict of interest.

\section*{Data availability}
The datasets generated for this paper are available at Zenodo. Separate submissions are made for the processed data resulting in radiographs with ground truth for object detection~\cite{Zeegers1}, as well as the unprocessed CT scan data for complete reproduction of the results in this paper~\cite{Zeegers2}.

\section*{Acknowledgements}
The authors acknowledge financial support from the Netherlands Organisation for Scientific Research (NWO), project number 639.073.506. D. M. Pelt is supported by The Netherlands Organisation for Scientific Research (NWO), project number 016.Veni.192.235. The authors also acknowledge TESCAN-XRE NV for their collaboration and support of the FleX-ray laboratory.

\appendices

\renewcommand\thefigure{\thesection\arabic{figure}}
\setcounter{figure}{0}

\section{Intensity value histograms}

\begin{figure}[H]
    \centering
    \subcaptionbox[c]{2D radiograph with foreign object on bottom left\label{fig:ProjectionDataLabObj3Appendix}}[0.16\textwidth]{
		\includegraphics[width=0.16\textwidth]{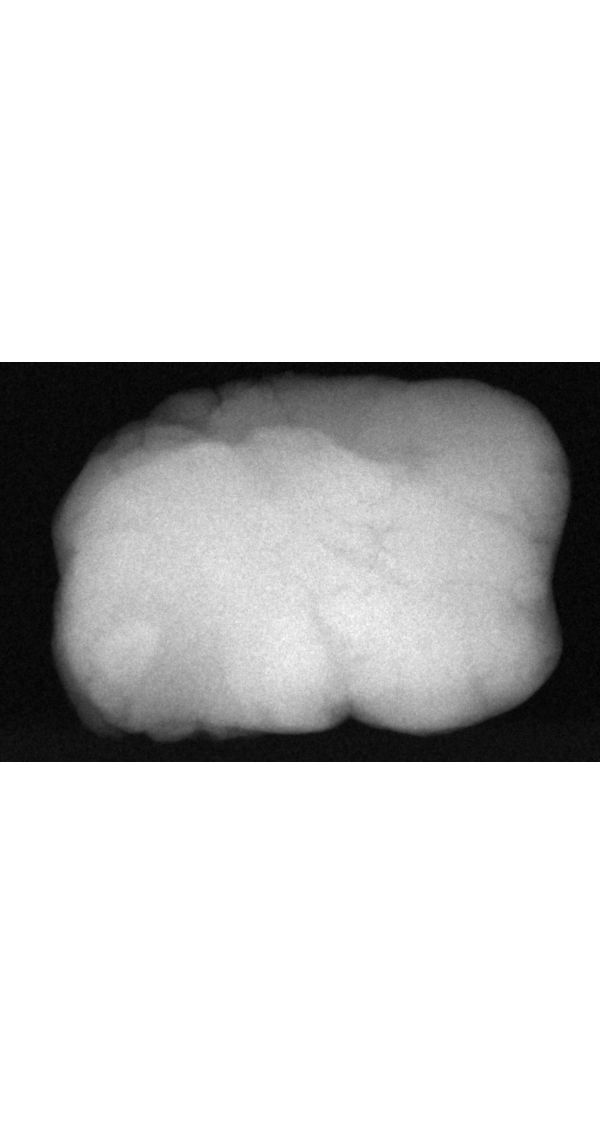}} \ \
    \subcaptionbox[c]{Intensity value distribution for the 2D radiograph\label{fig:Histogram1}}[0.4\textwidth]{
		\includegraphics[width=0.4\textwidth]{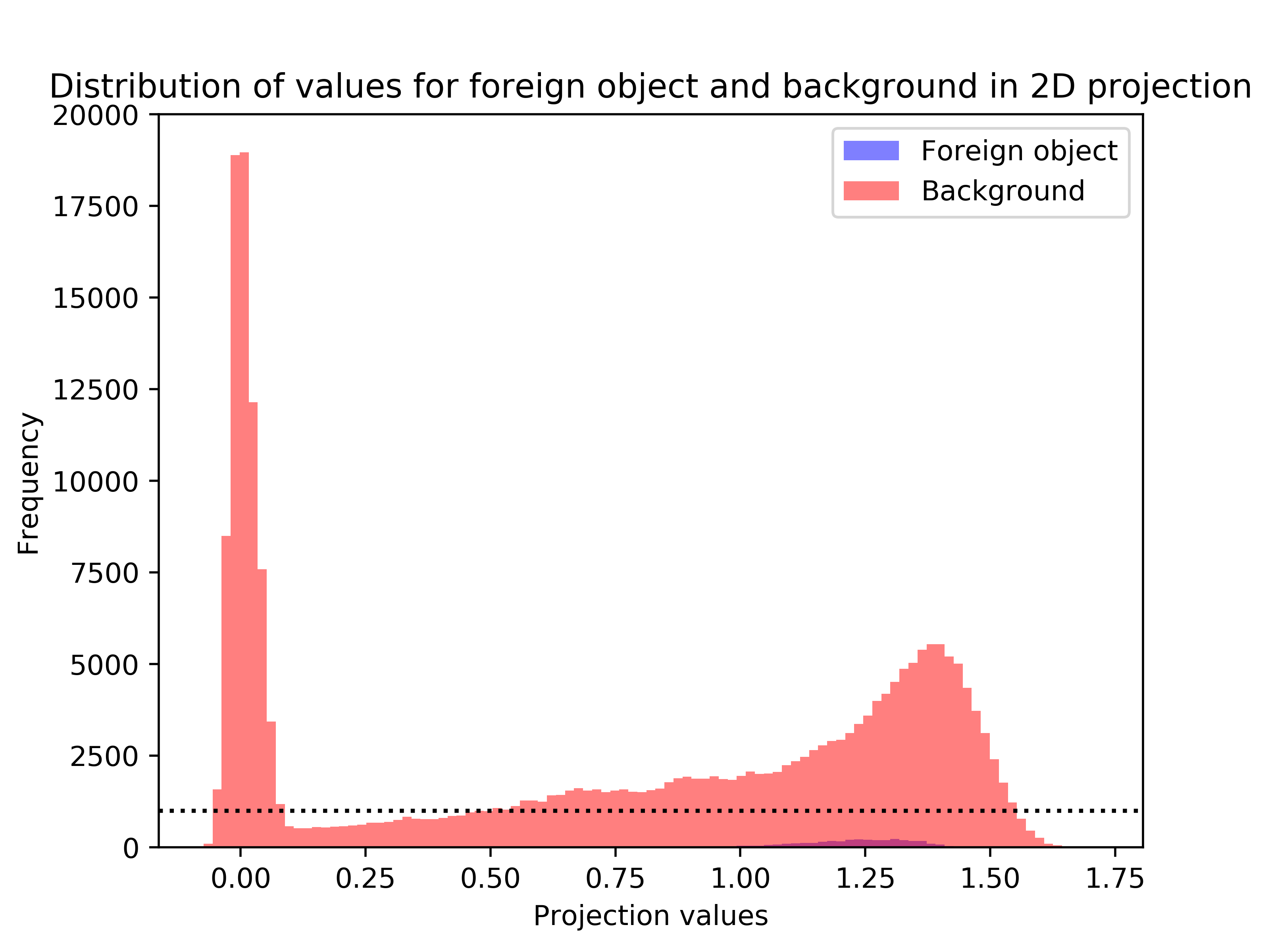}} \ \
	\subcaptionbox[c]{Intensity value distribution for the 2D radiograph (zoomed)\label{fig:Histogram1Zoomed}}[0.4\textwidth]{
		\includegraphics[width=0.4\textwidth]{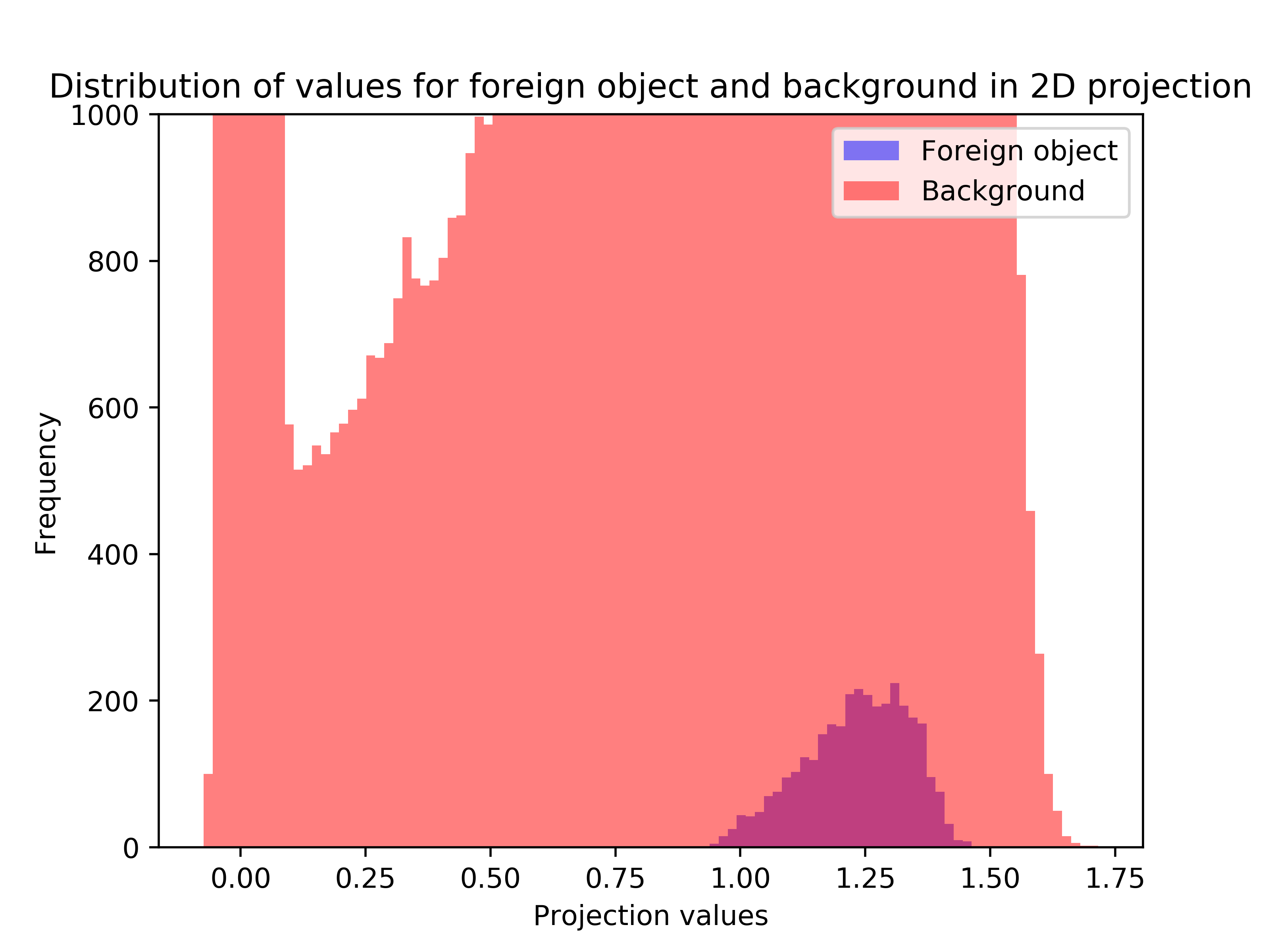}} \\[2mm]
	\subcaptionbox{Slice of the reconstructed 3D volume with foreign object on bottom left \label{fig:ReconstructionSlice}}[0.16\textwidth]{
		\includegraphics[width=0.16\textwidth]{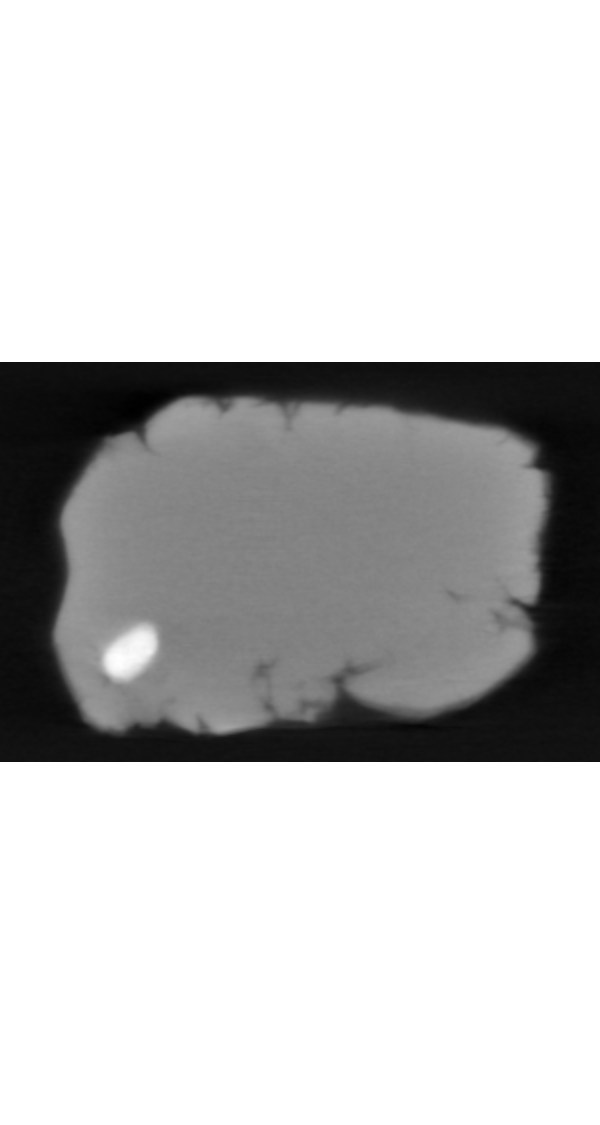}} \ \
    \subcaptionbox{Attenuation value distribution for the slice of the reconstructed 3D object \label{fig:Histogram2}}[0.4\textwidth]{
		\includegraphics[width=0.4\textwidth]{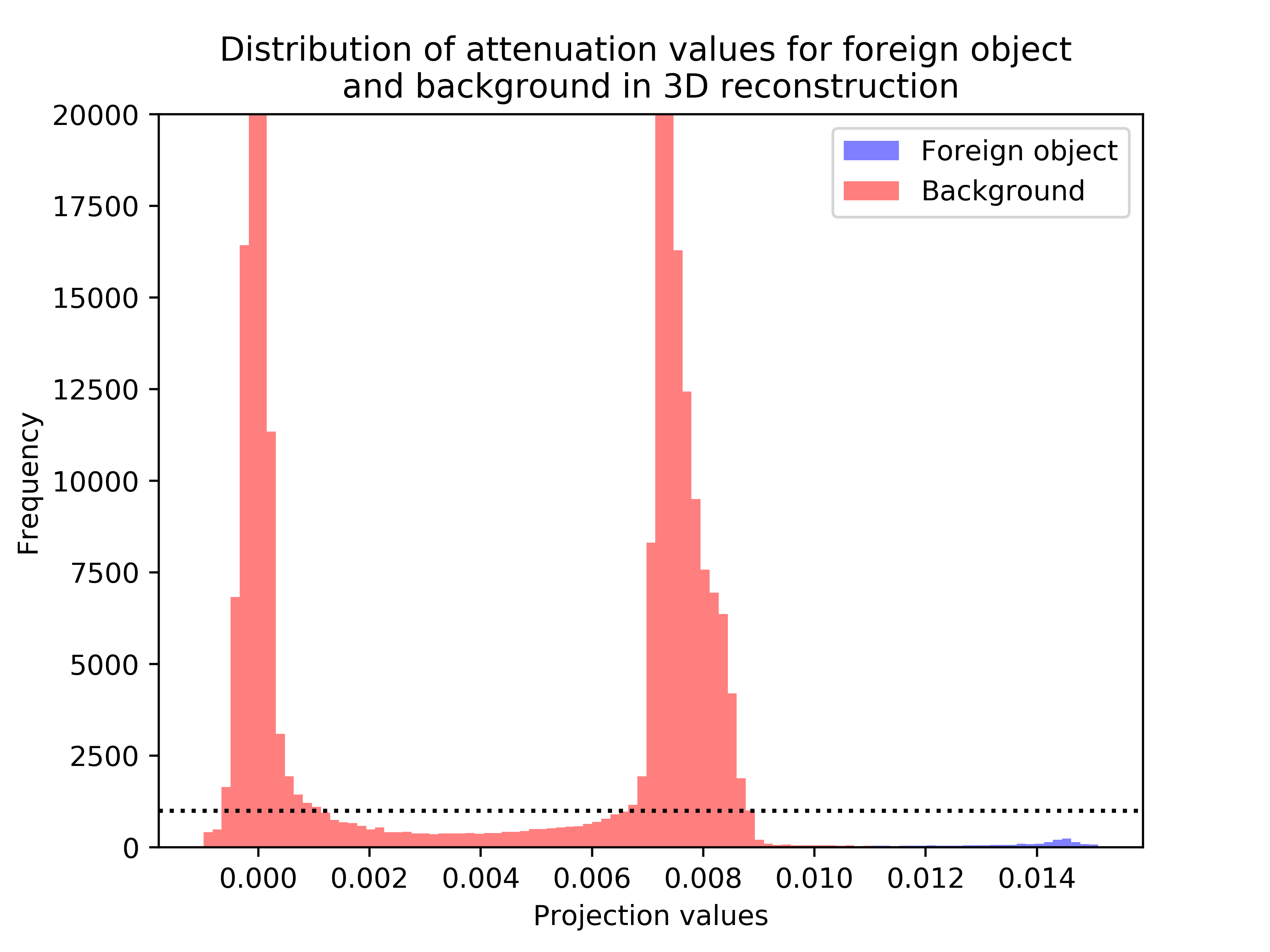}} \ \
	\subcaptionbox{Attenuation value distribution for the slice of the reconstructed 3D object (zoomed) \label{fig:Histogram2Zoomed}}[0.4\textwidth]{
		\includegraphics[width=0.4\textwidth]{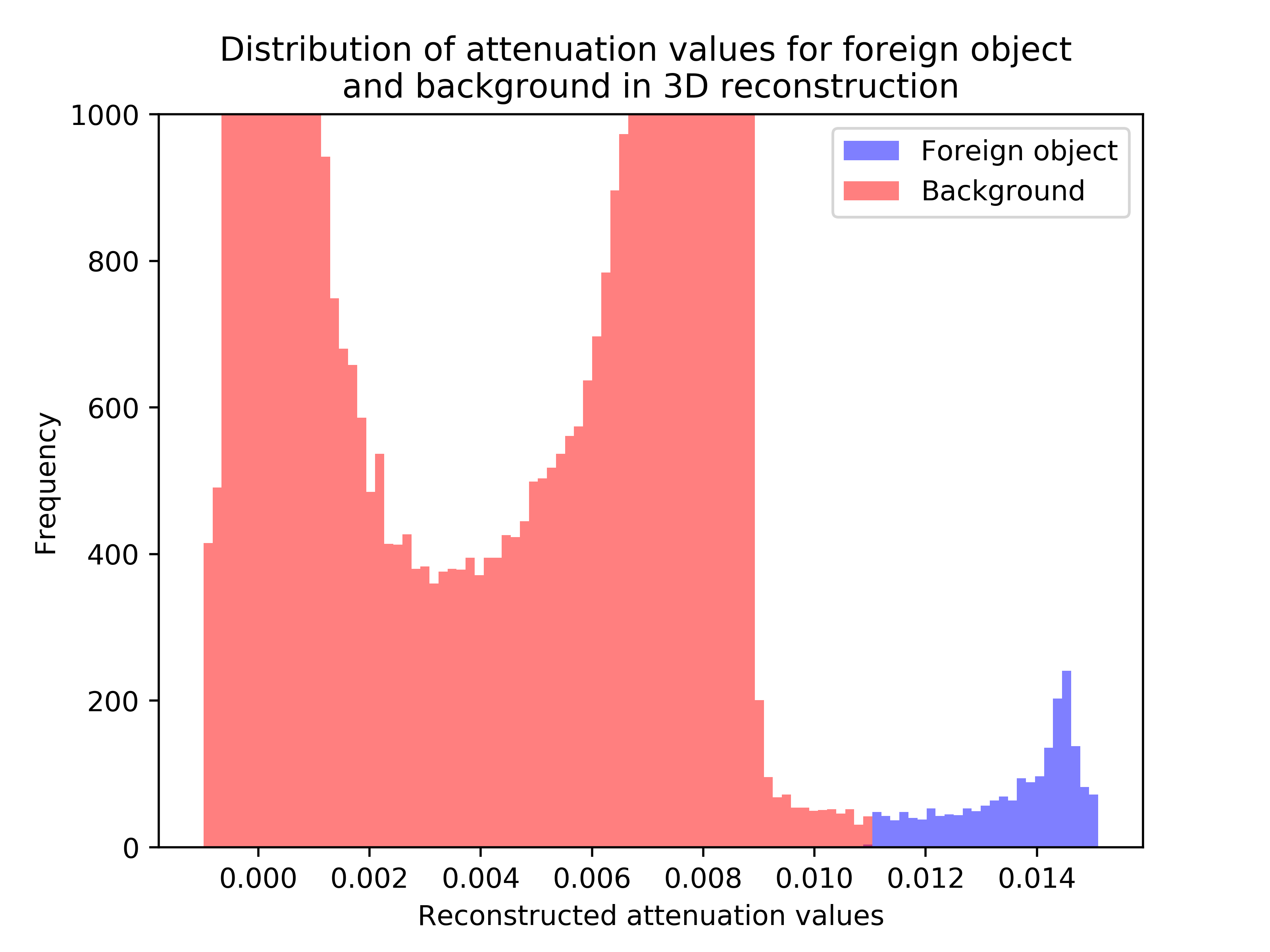}} \\[2mm]
	\caption{Radiograph of an object containing a foreign object (\textbf{a}) and a slice of the corresponding 3D reconstruction showing its attenuation values (\textbf{d}), indicating the difference in contrast. Additionally, histograms of intensity value distribution of the radiograph (\textbf{b}-\textbf{d}) and the attenuation value distribution of the slice of the reconstructed 3D object (\textbf{e}-\textbf{f}). In both cases, the histograms of the voxels or pixels of the foreign object are plotted separately from the other voxels or pixels. In the 3D volume, the foreign object is much easier to distinguish based on intensity values.}\label{fig:Separation}
\end{figure}

We compare the intensity distributions for radiographs and for a CT scan for an object in Figure~\ref{fig:Separation}, which shows a number of statistics about the pixel and voxel intensities for object 3 (Fig.~\ref{fig:ProjectionDataLab}). For both approaches, the intensity value distributions are plotted and separated into values of pixel or voxels that have been marked as foreign object by the thresholding method. The 3D case has a clear separation between foreign object and the base object based on attenuation, such that a simple global threshold based on Otsu's method \cite{Otsu} is sufficient to segment the foreign object. On the other hand, in the 2D radiograph case, the intensity values corresponding to the foreign object locations are similar to values of the base object.

\end{document}